\documentclass{article}

\usepackage{arxiv}

\usepackage[utf8]{inputenc} 
\usepackage[T1]{fontenc}    
\usepackage{hyperref}       
\usepackage{url}            
\usepackage{booktabs}       
\usepackage{amsfonts}       
\usepackage{nicefrac}       
\usepackage{microtype}      
\usepackage{lipsum}		
\usepackage{graphicx}
\usepackage{natbib}
\usepackage{doi}

\usepackage{amsmath}
\usepackage{amssymb}
\usepackage{mathtools}
\usepackage{amsthm}
\usepackage{bbm}
\usepackage{enumitem}
\usepackage{multirow}
\usepackage[capitalize,noabbrev]{cleveref}

\theoremstyle{plain}
\newtheorem{theorem}{Theorem}[section]

\newtheorem{claim}{Claim}
\newtheorem{hypothesis}{Hypothesis}
\theoremstyle{definition}
\newtheorem{definition}[theorem]{Definition}

\theoremstyle{remark}

\title{Revisiting Generalization Power of a DNN \\
in Terms of Symbolic Interactions}

\date{} 					

\author{  Lei Cheng,\\
    Shanghai Jiao Tong University\\
    \texttt{chenglei20020408@sjtu.edu.cn}\\
    \And
    Junpeng Zhang,\\
    Shanghai Jiao Tong University\\
    \texttt{zhangjp63@sjtu.edu.cn}\\
    \And
    Qihan Ren,\\
    Shanghai Jiao Tong University\\
    \texttt{renqihan@sjtu.edu.cn}\\
    \And
    Quanshi Zhang\thanks{Quanshi Zhang is the corresponding author. He is with the Department of Computer Science and Engineering, the John Hopcroft Center, at the Shanghai Jiao Tong University, China.}\\
    Shanghai Jiao Tong University\\
    \texttt{zqs1022@sjtu.edu.cn}
}

\renewcommand{\headeright}{}
\renewcommand{\undertitle}{}
\renewcommand{\shorttitle}{Revisiting Generalization Power of a DNN in Terms of Symbolic Interactions}

\hypersetup{
pdftitle={A template for the arxiv style},
pdfsubject={q-bio.NC, q-bio.QM},
pdfauthor={David S.~Hippocampus, Elias D.~Striatum},
pdfkeywords={First keyword, Second keyword, More},
}

\begin{document}
\maketitle

\begin{abstract}
    This paper aims to analyze the generalization power of deep neural networks (DNNs) from the perspective of interactions. Unlike previous analysis of a DNN's generalization power in a high-dimensional feature space, we find that the generalization power of a DNN can be explained as the generalization power of the interactions. We found that the generalizable interactions follow a decay-shaped distribution, while non-generalizable interactions follow a spindle-shaped distribution.  Furthermore, our theory can effectively disentangle these two types of interactions from a DNN. We have verified that our theory can well match real interactions in a DNN in experiments.
\end{abstract}


\section{Introduction}
Analyzing and quantifying the generalization power of deep neural networks (DNNs) is a crucial issue in deep learning. For example, numerous achievements have been made to analyze the generalization power in high-dimensional feature space.~\citep{petrini2022learning, boopathy2023model, dyballa2024separability,nikolikj2024generalization}

\begin{figure*}[ht]
	\centering
	\includegraphics[width=\linewidth]{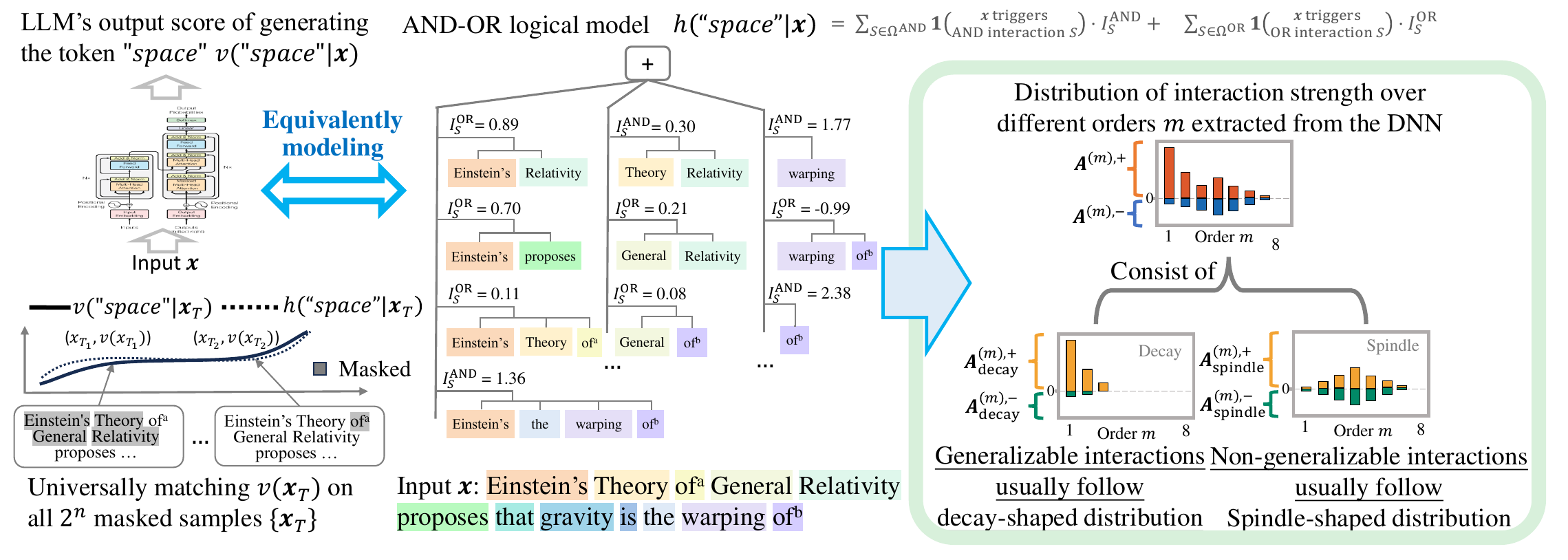}
	\caption{{(Left) It is proven that there exists a logical model consisting of AND-OR interactions, which can accurately predict all the DNN's outputs, when we augment the input by enumerating its all $2^n$ masked states. (Right) We have found that interactions in a DNN can be decomposed into a set of generalizable interactions following a decay-shaped distribution and a set of non-generalizable interactions following a spindle-shaped distribution.}}
	\label{fig:fig1}
\end{figure*}
{
In this paper, we revisit the generalization power of a DNN from a new perspective. Unlike analyzing a DNN's generalization power in the high-dimensional feature space, people usually have a more efficient way to evaluate the \textit{trustworthiness} of another person, \textit{i.e.,} directly understanding the internal logic behind his/her behavior.} \textit{However, this strategy has not been conducted on DNNs, because the lack of a sophisticated theoretical system to faithfully quantify/explain detailed logic (or inference patterns) encoded by a DNN has hampered all attempts {of using detailed inference patterns to analyze a DNN's generalization power.}}

\textbf{Background: faithful {disentanglement} of interaction patterns used by a DNN for inference.} The first challenge to the above strategy is the theoretically guaranteed faithfulness of inference patterns. To this end, the interaction theory with lots of theorems and empirical findings has been built up~\citep{Ren_2023_CVPR,pmlr-v202-li23at, ren2024where} to guarantee the faithfulness of {defining and disentangling} a DNN's inference patterns, and it has been widely applied to various scenarios~\citep{ren2021towards, wang2021unified, deng2022discovering, liu2023towards, zhou2024generalization, zhang2024two, ren2024towards}.

Specifically, as Figure~\ref{fig:fig1} shows, given a DNN and an input sample $\boldsymbol{x}$, it is proven that there exists a logical model consisting of AND-OR interaction patterns, {which can accurately predict the DNN's outputs on all $2^n$ masked states of the input sample.} Such a strong theorem enables us to roughly consider the set of \textit{AND-OR interactions} in the logical model as inference patterns equivalently encoded by the DNN. For example, in Figure~\ref{fig:fig1}, the logical model encodes an AND interaction between the tokens in {\small$S=\{Theory, Relativity\}$}. Only when all tokens in $S$ are present in the sentences, the interaction is activated and makes a numerical effect. In this way, no matter how we mask the input sentence, the logical model can always use the sum of all activated interactions to predict the DNN's output on this masked sample.

{
\textbf{Our work:} since the output of a DNN can be faithfully explained as the sum of all AND-OR interactions, \textbf{why not directly identify the generalizable interactions and non-generalizable interactions used by the DNN? In this way, we can roughly consider the generalization power of the entire DNN as the ensemble of the generalization power of all detailed AND-OR interaction patterns.}}

{
The perspective of interactions simplifies the understanding of the generalization power of the entire DNN, because the generalization power of each interaction can be quantified in a much more straightforward manner. Specifically, if an interaction that frequently appears in training samples also frequently appears in testing samples, then we consider this interaction is generalized to testing samples.}

Therefore, in this study, we aim to distinguish generalizable interactions and non-generalizable interactions. {Specifically, we find that the generalizable interactions and non-generalizable interactions follow two fully distinct distributions over different interaction complexities, where the complexity of an interaction $S$ is measured by the interaction order, \textit{i.e.}, the number of input variables in the set $S$,  $\textit{order}(S)=\vert S \vert$. Thus, we propose and further prove the following two hypothesises on the generalization power of interactions.}

(1) Generalizable interactions encoded by a DNN usually exhibit a \textbf{decay-shaped distribution} of interactions with increasing complexity (see Figure~\ref{fig:fig1}~(right)). It means that most generalizable interactions are of low orders, and very few generalizable interactions are of high orders.

{(2) Non-generalizable interactions exhibit a \textbf{spindle-shaped distribution} across different orders (complexities). It means that most non-generalizable interactions are of medium orders, and much less non-generalizable interactions are of extremely low orders or extremely high orders.}

The above hypothesises are motivated by the finding that low-order interactions are better generalized to testing samples than high-order interactions~\citep{zhou2024explaining}. Furthermore, \citet{ren2024towards, zhang2024two} have found that when the training of a DNN enters the overfitting phase, it mainly learns medium-order and high-order interactions.

We have conducted experiments to verify our hypothesis. Specifically, we discover before the overfitting phase, DNN mainly removes the interactions following a spindle-shaped distribution and converges to interactions following a decay-shaped distribution. In the overfitting phase, the DNN learns the interactions following a spindle-shaped distribution again. Furthermore, we find that injecting non-generalizable representations to a DNN generates new interactions following a spindle-shaped distribution.

{
\textbf{Disentangling the generalizable and non-generalizable interactions.} Based on the above theory, we propose a method to formulate and disentangle the spindle-shaped interactions and decay-shaped interactions. Experimental results show that our method accurately explained the two types of interactions in real DNNs.}

The contribution of this paper can be summarized as follows. In this study, we find that the generalizable interactions encoded by a DNN often follow a decay-shaped distribution, while the non-generalizable interactions often follow a spindle-shaped distribution. {Furthermore, we propose a method to formulate and disentangle interactions of a spindle-shaped distribution and interactions of a decay-shaped interactions from a DNN. Preliminary experiments have showed the effectiveness of our theory.}

{\section{Analyzing interactions of two distributions}
\subsection{Preliminaries: interactions represent primitive inference patterns used by a DNN}}
To analyze the DNN's generalization power from the perspective of detailed inference patterns, the first challenge is how to faithfully extract and quantify the inference patterns encoded by a DNN. 

As a typical theoretical guarantee for the faithfulness of explaining inference patterns used by a DNN, recent progresses \cite{Ren_2023_CVPR, pmlr-v202-li23at, ren2024where} have proved {that a DNN's all detailed inference patterns on a given input sample can be rigorously mimicked by a logical model consisting of AND-OR interactions.} Let us be given a DNN and an input sample $\boldsymbol{x}=[x_1, x_2, \ldots, x_n]$ consisting of $n$ input variables\footnote{{Please see Appendix~\ref{sec:apdx-detail-compute-interaction} for how to use a token embedding or an image patch as a single input variable.}}. {\small$N=\{ 1, 2, \ldots, n \}$} denotes the set of indices of the $n$ input variables. We use $v(\boldsymbol{x})\in\mathbb{R}$ to represent the scalar output of the DNN. Typically, we can follow~\cite{deng2022discovering} to set $v(\boldsymbol{x})$ as the scalar classification confidence\footnote{{Please see Appendix~\ref{sec:apdx-detail-compute-interaction} for the probability in the tasks of language generation and image classification.}} of the sample $\boldsymbol{x}$.
\begin{equation}\begin{aligned}
v(\boldsymbol{x}) \overset{\text{def}}{=} \text{log} \frac{p(y=y^{\text{truth}} \vert \boldsymbol{x})}{1 - p(y=y^{\text{truth}} \vert \boldsymbol{x})} \in \mathbb{R}.
\label{eq:harsanyi_interaction_and}
\end{aligned}\end{equation}
where $p(y=y^{\text{truth}} \vert \boldsymbol{x})$ represents the probability\footnotemark[3] of classifying the input sample $\boldsymbol{x}$ to the ground-truth category $y^{\text{truth}}$.

Theorem~\ref{them:matching} theoretically guarantees that given an input sample $\boldsymbol{x}$, we can construct the following AND-OR logical model to accurately predict the DNN's scalar outputs on all masked samples $\boldsymbol{x}_\text{mask}$\footnote{\label{footnote:mask}People usually mask the input variables $i$ by setting $\boldsymbol{x}_i$ to the baseline values $b_i$. {In image classification, we follow \citet{dabkowski2017real} to set {\small$b_i$} as the average color value across all pixels in all images. Given an LLM, we follow \cite{shen2023inference} to set {\small$b_i$} as a special token (\textit{i.e.,} [MASK] token).}} when we enumerate all $2^n$ possible masked states. 
\begin{equation}\begin{aligned}
h(\boldsymbol{x}_\text{mask}) \overset{\text{def}}{=}  &\sum\nolimits_{S\in \Omega^{\text{AND}}} \mathbbm{1}_{\text{AND}}(S \vert \boldsymbol{x}_\text{mask})  \cdot I_S^{\text{AND}} \\ + &\sum\nolimits_{S\in \Omega^{\text{OR}}} \mathbbm{1}_{\text{OR}}(S \vert \boldsymbol{x}_\text{mask})  \cdot I_S^{\text{OR}} + b
\label{eq:and_or_model}
\end{aligned}\end{equation}

{\small $\bullet$}\;\textbf{The trigger function $\mathbbm{1}_{\text{\rm AND}}(S \vert \boldsymbol{x}_\text{\rm mask})$ represents an AND interaction between a subset $S\subseteq N$ of input variables.} If all input variables in $S$ are present (not masked) in the masked sample $\boldsymbol{x_\text{\rm mask}}$, the AND trigger function is activated and returns 1. Otherwise, it returns 0. $I_S^{\text{AND}}$ is the scalar weight. The bias term $b$ is set to $v(\boldsymbol{x}_\emptyset)$, {\textit{i.e.}, the network output} when masking all input variables in $\boldsymbol{x}$. $\Omega^{\text{AND}}$ and $\Omega^{\text{OR}}$ represent the set of AND interactions and the set of OR interactions, respectively.

{\small $\bullet$}\;\textbf{The trigger function $\mathbbm{1}_{\text{\rm OR}}(S \vert \boldsymbol{x_\text{\rm mask}})$ represents an OR interaction {between} a subset $S\subseteq N$ of input variables.} If at least one input variable in $S$ is present {(not masked)} in $\boldsymbol{x_\text{mask}}$, then the OR trigger function is activated and returns 1. Otherwise, it returns 0. $I_S^{\text{OR}}$ is the scalar weight.

\begin{theorem}\label{them:matching}
\textbf{(Universal matching property, proved by~\citet{chen2024defining} and Appendix~\ref{proof:universal-matching})}
Given an input sample {$\boldsymbol{x}$}, if we set all weights {$\forall S\subseteq N$}, $I_S^{\text{AND}}= \sum\nolimits_{T\subseteq S} (-1)^{\vert S \vert - \vert T \vert}\cdot u^\text{AND}_T$ and $I_S^{\text{OR}}= -\sum\nolimits_{T\subseteq S} $ $ (-1)^{\vert S \vert - \vert T \vert}\cdot u^\text{OR}_{N\setminus T}$, then we have
\begin{equation}\begin{aligned}
    \forall\ T \subseteq N, \quad v(\boldsymbol{x}_T) = h(\boldsymbol{x}_T)
\end{aligned}\end{equation}
where $\boldsymbol{x}_T$ is the masked sample only containing the input variables in $T$. All other input variables in the set $N\backslash T$ are masked.\footnotemark[4] The overall network output {\small$v\boldsymbol(x_T)$} is decomposed, subject to $u^\text{AND}_T = 0.5\cdot v(\boldsymbol{x}_T) + \gamma_T$ and $u^\text{OR}_T = 0.5\cdot v(\boldsymbol{x}_T) - \gamma_T$. $\{\gamma_T\}$ is a set of learnable parameters to determine the decomposition (See Appendix~\ref{sec:apdx-optimize-pq} for the details). 
\end{theorem}
This theorem ensures that each logical model can predict DNN's outputs on an exponential number of masked samples. \textbf{This guarantees that we can roughly consider AND-OR interactions in the logical model as primitive inference patterns equivalently used by the DNN.}

{\textbf{Sparsity property} of interactions is another important guarantee for the faithfulness of interactions.} \citet{ren2024where} have proven that a well-trained DNN usually only encodes very sparse interactions.\footnote{The sparsity of interactions can be guaranteed by three common conditions for the DNN's smooth inferences on randomly masked samples. Detailed conditions are provided in the Appendix~\ref{sec:apdx-condition-for-sparsity}.} \textit{I.e.}, {we can learn sparse interactions by optimizing parameters $\{\gamma_T\}$ defined in Theorem~\ref{them:matching} via {\small$\min \sum_S \vert I_S^\text{AND}\vert+ \sum_S \vert I_S^\text{OR}\vert$}. In this way, it is proven that only $O(n^p)$ interactions have salient effects,where $p \in [1.5, 2]$. All other interactions have negligible effects $I_S^{\text{AND}}\approx 0, I_S^{\text{OR}}\approx 0$. In words, we can use very few salient interactions to construct $\Omega^{\text{AND}}$ and $\Omega^{\text{OR}}$ to approximate the DNN's outputs on all masked samples.}

\subsection{Two perspectives to define the generalization power of interactions}
\label{sec:two-perspectives}
Because the output of a DNN can be faithfully decomposed into effects of AND-OR interactions, we can explain the generalization power of a DNN as the ensemble of the generalization power of its compositional interactions. 

\textbf{The first perspective to understand the generalization power of an interaction.} Definition~\ref{def:generalization} shows a typical definition of the generalization power of interactions proposed by \citet{zhou2024explaining}. If an interaction, which frequently appears on the training samples, also frequently appears on the testing samples, then this interaction is considered to be generalized to testing samples; otherwise, not.

\begin{definition}
\label{def:generalization}
Given a set of AND interactions and a set of OR interactions, the generalization power of all these AND-OR interactions is defined as the Jaccard similarity $Sim(D_{\text{train}}, D_{\text{test}})$ between these interactions' distributions $D_{\text{train}}$\footnote{Given a set of $k$ AND interactions and $l$ OR interactions extracted from the sample $\boldsymbol{x}$, let us vectorize all interactions into a vector with non-negative elements as $\textbf{I}(\boldsymbol{x})=\text{max}\{[I_{S_1}^{\text{AND}},-I_{S_1}^{\text{AND}},I_{S_2}^{\text{AND}},-I_{S_2}^{\text{AND}},\ldots,I_{S_k}^{\text{AND}},-I_{S_k}^{\text{AND}},I_{S_1}^{\text{OR}},-I_{S_1}^{\text{OR}},$ $\ldots,I_{S_l}^{\text{OR}},-I_{S_l}^{\text{OR}}]^T,0\}$. Then, the distribution of these interactions over training samples is measured as the average effect of these interactions over training samples, \textit{i.e.}, $D_{\text{train}}=\mathbb{E}_{\boldsymbol{x}\in \text{train set}} \textbf{I}(\boldsymbol{x})$. Similarly, the distribution over testing samples is given as $D_{\text{test}}=\mathbb{E}_{\boldsymbol{x}\in \text{test set}} \textbf{I}(\boldsymbol{x})$.} over the training set and their distributions $D_{\text{test}}$\footnotemark[6] over the testing set, \textit{i.e.}, $Sim(D_{\text{train}}, D_{\text{test}}) = \frac{\Vert \min\{D_{\text{train}}, D_{\text{test}}\} \Vert_1}{\Vert \max\{D_{\text{train}}, D_{\text{test}}\} \Vert_1}$, where $\Vert \cdot \Vert_1$ denotes the $\ell_1$ norm.
\end{definition}

In this way, \citet{zhou2024explaining} have found the following empirical findings on interactions of each order. The order (or complexity) of an interaction is defined as the number of input variables in the set $S$,~\textit{i.e.,}  $\textit{order}(S)=\vert S \vert$. The experimental verification of the following claim is provided in Appendix~\ref{sec:apdx-verify-generalization}.
\begin{claim}[\textbf{Generalization power of interactions of different orders.}]
\label{claim:generalization}
The generalization power of high-order (complex) interactions is lower than that of low-order (simple) interactions, \textit{i.e.}, low-order interactions have a higher value of $\text{Sim}(D_{\text{train}}, D_{\text{test}})$ than high-order interactions.
\end{claim}

\textbf{The second perspective to understand the generalization power of an interaction.} In this study, we find that the above claim does not explain the fundamental ways for a DNN to encode generalizable interactions and non-generalizable interactions. Instead, we propose and later verify the following hypothesis on how a DNN encodes generalizable interactions and non-generalizable interactions.
\begin{hypothesis}
\label{hypothesis:generalization}
{Generalizable interactions encoded by a DNN usually follow a decay-shaped distribution over different orders, while non-generalizable interactions encoded in the DNN usually follow a spindle-shaped distribution over different orders.}
\end{hypothesis}
We follow \citet{ren2024towards} to represent the distribution of interactions mentioned in Hypothesis~\ref{hypothesis:generalization}. Given all interactions of the $m$-th order $\{ S\subseteq N : \vert S\vert=m\}$, we compute the strength of all positive interactions $\textbf{A}^{(m),+}$ and all negative interactions $\textbf{A}^{(m),-}$ as follows:
\begin{equation}\begin{aligned}
\label{eq:A}
\textbf{A}^{(m),+}=\sum_{S:\vert S \vert = m} \max\{I_S^{\text{AND}}, 0\} + \max\{I_S^{\text{OR}}, 0\} \\
\textbf{A}^{(m),-}=-\sum_{S:\vert S \vert = m} \min\{I_S^{\text{AND}}, 0\} + \min\{I_S^{\text{OR}}, 0\}
\end{aligned}\end{equation}
In Figures~\ref{fig:fig1}, \ref{fig:stage} and \ref{fig:verify_disentangle}, we visualize $\textbf{A}^{(m),+}$ and $\textbf{A}^{(m),+}$ over different orders to represent the decay-shaped distribution of interactions and the spindle-shaped distribution of interactions in Hypothesis~\ref{hypothesis:generalization}.

In order to verify Hypothesis~\ref{hypothesis:generalization}, we conducted experiments in Sections~\ref{sec:two_stage} and \ref{sec:generalization_power} to compare (1) the distribution of interactions extracted from a well-trained DNN with (2) the distribution when we revise this DNN towards overfitting. Then, Hypothesis~\ref{hypothesis:generalization} can be verified if we find that (1) the well-trained DNN usually encodes decay-shaped interactions; (2) interactions of spindle-shaped distributions newly emerge when a well-trained DNN is revised to encode additional non-generalizable features. 

\subsubsection{Verifying the hypothesis in the two-stage dynamics of interactions}
\label{sec:two_stage}
Our hypothesis is grounded on the two-stage dynamics of interactions during the learning process of a DNN, which was found and proven by \citet{ren2024towards,zhang2024two}. Specifically, people can use the gap between the training loss and the testing loss to divide the entire training process into \textit{a normal learning phase} and \textit{an overfitting phase}. In the normal learning phase, the loss gap remains close to zero, and the DNN mainly encodes low-order interactions. In the overfitting phase, the loss gap suddenly begins to increase, and the DNN learns interactions of increasing orders.

In this subsection, we go beyond interactions of a single order and analyze the distribution of interactions in the two-stage dynamics, so as to verify Hypothesis~\ref{hypothesis:generalization}.

{\small $\bullet$}\;  \textbf{Stage 1: Normal learning phase.} We discover that the normal learning phase mainly pushes the DNN to encode interactions that follow a decay-shaped distribution over different orders. This discovery partially verifies Hypothesis~\ref{hypothesis:generalization}.

\begin{figure*}[t]
	\flushright  
	\includegraphics[width=\linewidth]{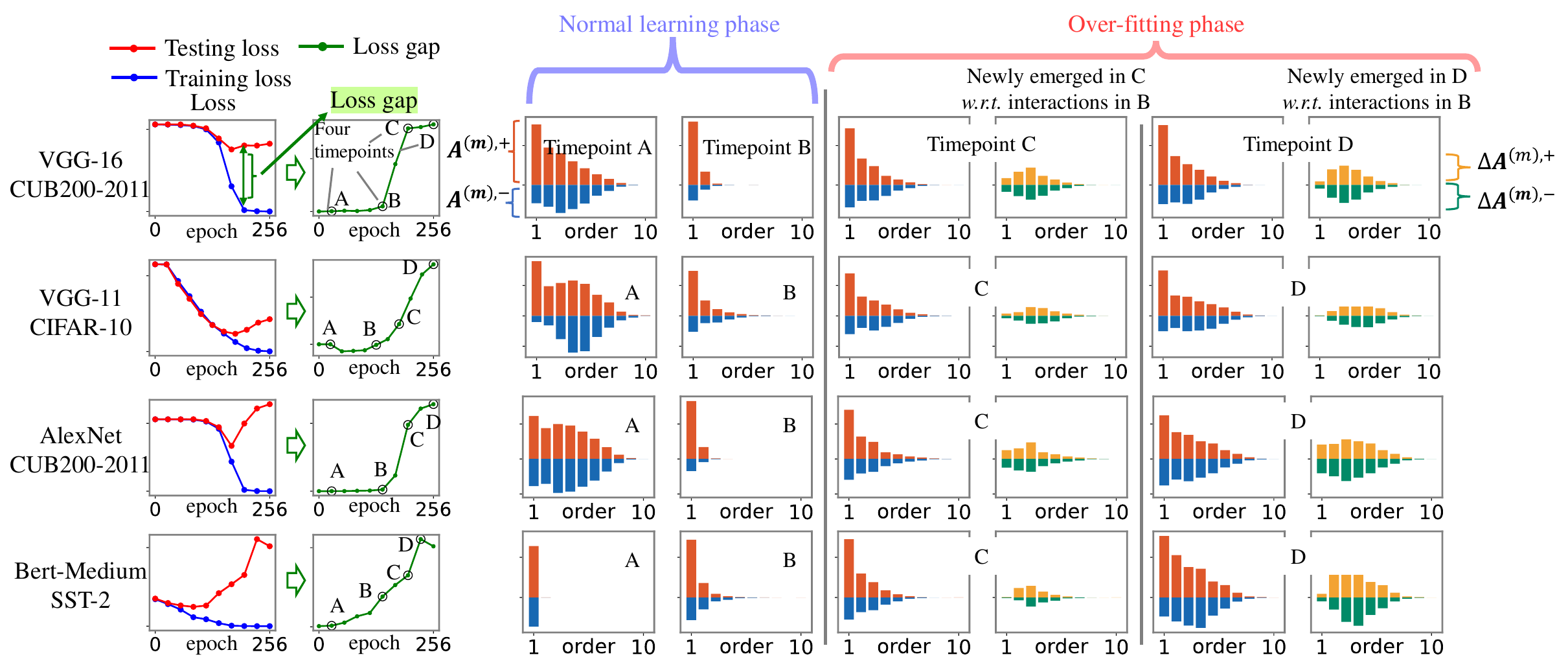}
	\caption{The two-stage dynamics of interactions in the learning of a DNN. In the first stage, noise interactions generated (at Timepoint A) by randomly initialized parameters were gradually removed (at Timepoint B), and the DNN mainly encoded interactions of a decay-shaped distribution. In the second stage, the newly emerged/learned interactions $\Delta\textbf{A}^{(m), +}$ and $\Delta\textbf{A}^{(m), -}$ follow a spindle-shaped distribution (see Timepoints C and D).}
	\label{fig:stage}
\end{figure*}
\textbf{\textit{Experiments.}} We conducted experiments to illustrate the distributions of interactions over different orders in the normal learning phase. We trained VGG-11, VGG-16~\citep{simonyan2014very}, AlexNet~\cite{krizhevsky2012imagenet} on the CIFAR-10 datasets~\citep{krizhevsky2009learning}, CUB200-2011 datasets~\cite{cub2011} and Tiny-ImageNet datasets~\cite{le2015tiny}, respectively. We also trained BERT-Medium models~\cite{devlin-etal-2019-bert} on the SST-2 datasets~\cite{socher2013recursive}. For image data, we followed \citet{ren2024towards} to set a random set of ten image patches as input variables. For natural language data, we set all embeddings related to a word as an input variable. Please see Appendix~\ref{sec:apdx-detail-compute-interaction} for the details. We selected two sets of salient interactions, and obtained $\Omega^\text{AND} = \{S \subseteq N: \vert I^\text{AND}_S \vert > \tau\}$ and $\Omega^\text{OR} = \{S \subseteq N: \vert I^\text{OR}_S \vert > \tau\}\ s.t.\ \tau=0.02 \cdot \mathbb{E}_{\boldsymbol{x}}\left[v(\boldsymbol{x})-v(\boldsymbol{x_\emptyset}) \right]$. Then, we measured the distributions of salient interactions in $\Omega^\text{AND}$ and $\Omega^\text{OR}$ by following Equation~(\ref{eq:A}).

Figure~\ref{fig:stage} shows the average distributions of $\textbf{A}^{(m),+}$ and the average distribution of  $\textbf{A}^{(m),-}$ of $m$-order interactions\footnote{We visualized the normalized strength of interactions for both clarity and fair comparison. Please see Appendix~\ref{sec:apdx-normalization} for the details.} over different samples. We found
that in the normal learning phase, a few noise interactions (following spindle-shaped distribution) generated by randomly initialized parameters were gradually removed. At the end of the normal learning phase, the DNN was usually considered well-trained, and it mainly encoded interactions of a decay-shaped distribution. These findings partially verified Hypothesis~\ref{hypothesis:generalization}, that the generalizable representations in DNNs are encoded as interactions of decay-shaped distributions.

{\small $\bullet$}\;  \textbf{Stage 2: Overfitting phase.} We discover that the overfitting phase usually causes the DNN to encode interactions of spindle-shaped distributions over different orders. {This partially verifies Hypothesis~\ref{hypothesis:generalization}.}

Let $\textbf{A}^{(m),+}$ and $\textbf{A}^{(m),-}$ denote the average distribution of interactions\footnotemark[7] extracted from the DNN at the end of the normal learning phase. $\textbf{A}_{\text{overfitting}}^{(m),+}$ and $\textbf{A}_{\text{overfitting}}^{(m),-}$ denote the average distribution of interactions extracted from the DNN in the overfitting phase. Then, we computed $\Delta \textbf{A}^{(m), +} = \left\vert \textbf{A}^{(m), +}_\text{overfitting} - \textbf{A}^{(m), +}\right\vert$ and $\Delta \textbf{A}^{(m), -} = \left\vert \textbf{A}^{(m), -}_\text{overfitting} - \textbf{A}^{(m), -}\right\vert$ to represented the distribution of the newly emerged interactions in the overfitting phase.

\textbf{\textit{Experiments.}} We continued training the DNNs towards overfitting. Figure~\ref{fig:stage} shows the distribution of newly emerged interactions $\Delta \textbf{A}^{(m), +}$ and $\Delta \textbf{A}^{(m), -}$ over different orders $m$ across different samples in the overfitting phase. All these distributions followed spindle-shaped distributions. This partially verified Hypothesis~\ref{hypothesis:generalization}, that the non-generalizable representations in DNNs are encoded as interactions of spindle-shaped distributions.

\subsubsection{verifying the spindle-shaped distribution of non-generalizable interactions}
\label{sec:generalization_power}
In this subsection, we revised a well-trained DNN by injecting non-generalizable representations into it to to obtain a DNN with non-generalizable features. In this way, \textit{we could examine whether interactions newly emerged in the revised DNN followed a spindle-shaped distribution.} {This experimental setting was also supported by the finding in \citep{zhang2024two}, \textit{i.e.}, the spindle-shaped distributions could also be derived by considering all newly emerged interactions acted like random noises on most training samples (in fact, they were learned to fit very few hard samples during the overfitting phase). Appendix~\ref{sec:apdx-discuss-adding-noise} provides further discussions on mathematical evidences of adding noises to generate non-generalizable representations.}

Therefore, in this subsection, we conducted two experiments to examine the newly emerged interactions, in which we revised the well-trained DNN from two perspectives.

In the first experiment, we revised the well-trained DNN by adding Gaussian noises $\epsilon \sim \mathcal{N}(0, \sigma^2)$ to the network parameters of a well-trained DNN. Then, we computed the interactions in the well-trained DNN, denoted by $I_S^{\text{AND}}$ and $I_S^{\text{OR}}$, and the interactions in the revised DNN, denoted by $I_{S, \text{noise}}^{\text{AND}}$ and $I_{S, \text{noise}}^{\text{OR}}$. Thus, $\Delta I_S^{\text{AND}} = I_S^{\text{AND}} - I_{S, \text{noise}}^{\text{AND}}$ and $\Delta I_S^{\text{OR}} = I_S^{\text{OR}} - I_{S, \text{noise}}^{\text{OR}}$ represented the interaction effects newly emerged in the revised DNN. We measured the distribution of newly emerged interactions in the revised DNN  $\Delta \textbf{A}^{(m), +}$ and  $\Delta \textbf{A}^{(m), -}$ by following Equation~(\ref{eq:A}).
\begin{equation}\begin{aligned}
    \Delta \textbf{A}^{(m), +} = \sum_{S:\vert S \vert = m} \max\{\Delta I_S^{\text{AND}}, 0\} + \max\{\Delta I_S^{\text{OR}}, 0\} \\
    \Delta \textbf{A}^{(m), -} = -\sum_{S:\vert S \vert = m} \min\{\Delta I_S^{\text{AND}}, 0\} + \min\{\Delta I_S^{\text{OR}}, 0\}
\end{aligned}\end{equation}
In the second experiment, given a specific sample $\boldsymbol{x}$, we revised this sample by adding adversarial perturbations\footnote{We applied the the Fast Gradient Sign Method (FGSM)~\citep{goodfellow2014explaining} to generate the adversarial perturbations $\boldsymbol{x}_\text{noise} = \boldsymbol{x} + \sigma \cdot \text{sign}(\nabla_{\boldsymbol{x}} v(\boldsymbol{x}))$, where $\sigma$ is the perturbation magnitude.} and generate $\boldsymbol{x}_\text{noise}$, as the injection of non-generalizable representations. Then, we used the same setting in the first experiment to measure the distribution of interactions newly emerged in the revised sample.

To this end, we conducted experiments on different DNNs, including ResNet-20~\citep{he2016deep}, VGG-11 and VGG-16 on various datasets, including MNIST~\citep{lecun1998gradient}, CIFAR-10 and CUB200-2011, respectively. 

\begin{figure*}[t!]
	\centering
	\includegraphics[width=\linewidth]{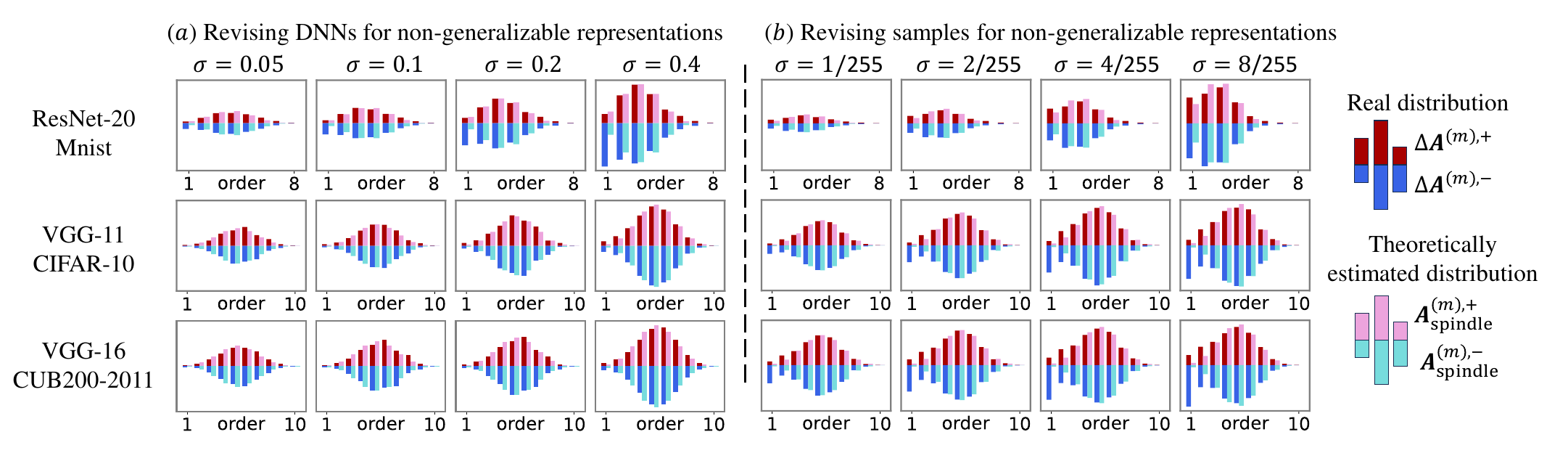}
	\caption{{The distributions of newly emerged interactions ($\Delta \textbf{A}^{(m), +}, \Delta \textbf{A}^{(m), -}$). All newly emerged interactions followed a spindle-shaped distribution. The magnitude of newly emerged interactions increased along with the increase of the injected non-generalizable representations.  The theoretical estimated  distributions of interactions $\{\textbf{A}^{(m),+}_{\text{Spindle}}, \textbf{A}^{(m),-}_{\text{Spindle}}\}$ in Equation~(\ref{eq:spindle}) well  matched the true distributions of newly-emerged interactions.}}
    \label{fig:reviseDNN}
\end{figure*}

{
Figure~\ref{fig:reviseDNN} shows the average distribution of interactions newly emerged in different samples in the above two experiments}. All newly emerged interactions followed a spindle-shaped distribution. Particularly, the significance of spindle-shaped distributions significantly increased when we added stronger noises. These experiments partially verified the hypothesis that non-generalizable interactions followed a spindle-shaped distribution.

\subsection{Modeling two distributions in the hypothesis}
\label{sec:modeling}
In this subsection, we derive the analytic solutions of the decay-shaped distribution of generalizable interactions and the spindle-shaped distribution of non-generalizable interactions, {in order to} better explain the observations in Sections~\ref{sec:two_stage} and \ref{sec:generalization_power}. Specifically, we propose two parameterized models to represent the above two distributions.

\subsubsection{Modeling the spindle-shaped distribution}
\label{sec:modeling_spindle}
Given a DNN with randomly initialized parameters, all AND-OR interactions in the DNN represent fully non-generalizable noises. Then, \citet{zhang2024two} have found that the effect of each non-generalizable interaction actually follows a Gaussian distribution.

In this way, we prove that these interactions follow a specific binomial distribution, \textit{i.e.}, $\textbf{A}^{(m), +} \propto \binom{n}{m}, \textbf{A}^{(m), -} \propto \binom{n}{m}$, where $\binom{n}{m} = \frac{n!}{m!(n-m)!}$ represents the combination number of $m$ input variables in $n$ input variables. Then, let us shift our attention from a fully initialized DNN to a trained DNN, in which not all input variables interact with each other. {Thus, we bring in a scaling coefficient $\alpha$ to formulate the distribution of interactions} $\textbf{A}^{(m),+}_{\text{spindle}}$ and $\textbf{A}^{(m),-}_{\text{spindle}}$.
\begin{equation}\begin{aligned}
    \label{eq:spindle}
    \textbf{A}^{(m), +}_{\text{spindle}}(\alpha, \beta) = 
    \beta \cdot \frac{\Gamma(\alpha n+1)}{\Gamma(m+1)\cdot\Gamma(\alpha n-m+1)} \\
    \textbf{A}^{(m), -}_{\text{spindle}}(\alpha, \beta) =
    \beta \cdot \frac{\Gamma(\alpha n+1)}{\Gamma(m+1)\cdot\Gamma(\alpha n-m+1)}
\end{aligned}\end{equation}
{where gamma function $\Gamma(\cdot)$ extends the binomial distribution $\binom{\alpha n}{m}$ to the domain of real numbers $\alpha n \in \mathbb{R}$. Specifically, the gamma function $\Gamma(\cdot)$ extends the factorial function to the real number field, \textit{i.e.}, $\Gamma(k) = (k+1)!$. $\beta$ is a coefficient for the magnitude of interactions.}

Despite the above simplifying settings, later experiments in Section~\ref{sec:experiments} showed that our theory accurately matched the real distribution of interactions in various DNNs (See Figure~\ref{fig:reviseDNN}).

\subsubsection{Modeling the decay-shaped distribution} 
{In this subsection, we use the optimal interactions under weight uncertainty proven by \citet{ren2024towards} to formulate the decay-shaped distribution of interactions.}

{
\begin{theorem}
\label{theorem:overfitting}(\textbf{Experimental verification is shown in Appendix~\ref{sec:apdx-exp-overfitting}}\footnote{Although the above theorem is difficult to understand, lots of experiments in \cite{ren2024towards} and Appendix~\ref{sec:apdx-exp-overfitting} have shown that Theorem~\ref{theorem:overfitting} has well predicted the dynamics of the interaction distribution before the overfitting of the DNN.})
Let $\textbf{I}^*=\left[I_{S_1}^*, I_{S_2}^*, \ldots, I_{S_{2^n}}^*\right]$ denote the effects of all $2^n$ interactions extracted from a fully converged DNN, which is probably overfitted. Then, injecting uncertainty/noises of the magnitude of $\delta$ to data during the training process would further reduce high-order interactions and yield the following set of interactions.
\begin{equation}\begin{aligned}
    \label{eq:predict_I}
    \hat{\textbf{I}}(\delta) = M(\delta)\  \textbf{I}^*
\end{aligned}\end{equation}
{where $M(\delta)\in \mathbb{R}^{2^n \times 2^n}$ is a matrix. The detail formula of $M$ is shown} in Appendix~\ref{sec:apdx-exp-overfitting} due to the limit of page length. The larger value of $\delta$ more penalizes the strength of high-order interactions.
\end{theorem}}
Theorem~\ref{theorem:overfitting} shows that the ubiquitous uncertainty in network parameters and training data will eliminate the mutually counteracting high-order interactions, thereby avoiding overfitting. In this way, we formulate the strength of positive interaction effect and that of negative interaction effect of each $m$-th order, as follows.
\begin{equation}\begin{aligned}
    \label{eq:decay}
    \textbf{A}^{(m), +}_{\text{decay}}(\delta, \gamma) = \gamma\sum_{S:\vert S \vert = m} \max\{\hat{I}_S^{\text{AND}}(\delta), 0\} + \max\{\hat{I}_S^{\text{OR}}(\delta), 0\} \\ 
    \textbf{A}^{(m), -}_{\text{decay}}(\delta, \gamma) = -\gamma\sum_{S:\vert S \vert = m} \min\{\hat{I}_S^{\text{AND}}(\delta), 0\} + \min\{\hat{I}_S^{\text{OR}}(\delta), 0\}
\end{aligned}\end{equation}
where $\gamma$ is a coefficient for the magnitude of interactions.

\textbf{Why do we choose Theorem~\ref{them:matching} to formulate the decay-shaped distribution?} Equation~(\ref{eq:decay}) does not describe the strength of each individual interaction, but it predicts the distribution of interactions over different orders\footnotemark[9]. The matrix $M(\delta)$ significantly weakens high-order interactions (see Appendix~\ref{sec:apdx-exp-overfitting}). Thus, we can consider that Theorem~\ref{theorem:overfitting} reflects an intrinsic nature of a DNN, \textit{i.e.}, the data uncertainty makes the training process weaken high-order interactions of a DNN, and ensures the generalization power of the DNN.

More importantly, despite the complexity of Theorem~\ref{theorem:overfitting}, later experiments will demonstrate the success of our choice, \textit{i.e.}, using the penalized interactions under noises to mimic the decay-shaped distribution of interactions.

\subsubsection{Disentangling the generalizable and non-generalizable interactions.} We  propose a method to disentangle the two distributions of interactions from a real DNN. Although the distributions in Equation~(\ref{eq:decay}) and Equation~(\ref{eq:spindle}) are derived on some simplifying assumptions and heuristic settings, experiments in Section~\ref{sec:experiments} will verify the effectiveness of our theory.
Given a DNN, let $\textbf{A}^{(m), +}$ and $\textbf{A}^{(m), -}$ denote the sum absolute strength of $m$-order positive interactions and negative interactions, respectively. Then, the distributions of generalizable interactions $\textbf{A}^{(m), +}_{\text{decay}}, \textbf{A}^{(m), -}_{\text{decay}}$ and non-generalizable interactions $\textbf{A}^{(m), +}_{\text{spindle}}, \textbf{A}^{(m), -}_{\text{spindle}}$ can be disentangled as the optimization of the following parameter model:
\begin{equation}\begin{aligned}
    \label{eq:disentangle}
    \min_{\alpha, \beta; \delta, \gamma} &\sum_{m=1}^n (\textbf{A}^{(m), +} - \textbf{A}_{\text{theory}}^{(m), +})^2 + (\textbf{A}^{(m), -} - \textbf{A}_{\text{theory}}^{(m), -} )^2 \\
    \text{s.t.} &\quad \textbf{A}_{\text{theory}}^{(m), +} = \textbf{A}^{(m), +}_{\text{decay}}(\delta, \gamma) + \textbf{A}^{(m), +}_{\text{spindle}}(\alpha, \beta) \\
    &\quad \textbf{A}_{\text{theory}}^{(m), -} = \textbf{A}^{(m), -}_{\text{decay}}(\delta, \gamma) + \textbf{A}^{(m), -}_{\text{spindle}}(\alpha, \beta)
\end{aligned}\end{equation}

\section{Experimental verification}
\label{sec:experiments}

\textbf{Fitness to the distribution of non-generalizable interactions.} We conducted experiments to verify whether the proposed formulation for the spindle-shaped distribution in Equation~(\ref{eq:spindle}) really represented the true distribution of non-generalizable interactions in a real DNN. Specifically, let us revisit the two experiments in Section~\ref{sec:generalization_power}, in which we proposed two ways to inject non-generalizable representations to a well-trained DNN. Thus, let us be given the distribution of interactions newly emerged in the revised DNN (or the revised samples), denoted by $\Delta \textbf{A}^{(m), +}, \Delta \textbf{A}^{(m), -}$, which were measured in two experiments in Section~\ref{sec:generalization_power}. We further used the derived distribution $\textbf{A}^{(m), +}_{\text{spindle}}, \textbf{A}^{(m), -}_{\text{spindle}}$ in Equation~(\ref{eq:spindle}) to match these interactions, \textit{i.e.}, computing parameters $\alpha$ and $\beta$ that best matched the distribution of these interactions. Figure~\ref{fig:reviseDNN} shows that the theoretical distributions $\textbf{A}^{(m), +}_{\text{spindle}}, \textbf{A}^{(m), -}_{\text{spindle}}$ well matched the true distribution of non-generalizable interactions in a real DNN, which partially proved the faithfulness of our theory.

\begin{figure*}[t]
    \centering
    \includegraphics[width=\linewidth]{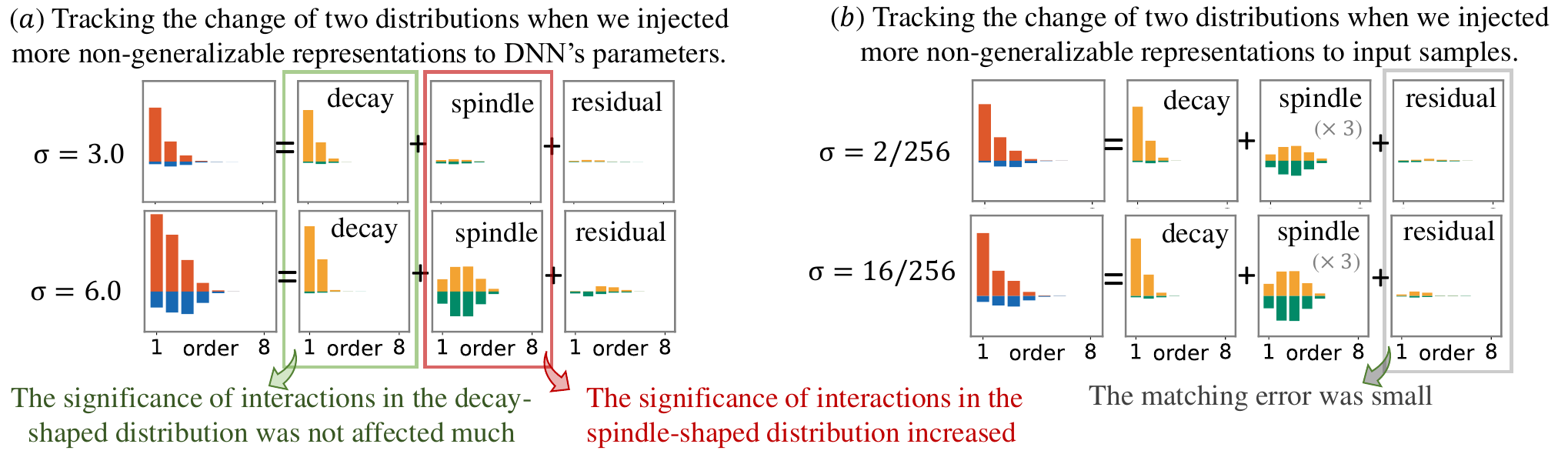}
    \caption
    {{The decay-shaped distribution of generalizable interactions ({\small$\textbf{A}^{(m),+}_{\text{decay}},\textbf{A}^{(m),-}_{\text{decay}}$}) and the spindle-shaped distribution of non-generalizable interactions ({\small$\textbf{A}^{(m),+}_{\text{spindle}}, \textbf{A}^{(m),-}_{\text{spindle}}$}) disentangled by our method. The significance of interactions in the spindle-shaped distribution increased when we injected more non-generalizable representations to the DNN, but the significance of interactions in a decay-shaped distribution was not affected much. It verified the faithfulness of our method. Please see  Appendix~\ref{sec:more-results-fig4} for more results of values {\small$\sigma$}.}}
    \label{fig:verify_disentangle}
\end{figure*}
\textbf{Faithfulness of simultaneously disentangling the distribution of generalizable interactions and that of non-generalizable interactions.} In this experiment, we followed Equation~(\ref{eq:disentangle}) to simultaneously disentangle the distribution of generalizable interactions and that of non-generalizable interactions from a given DNN. We evaluated the faithfulness of the disentangled two distributions in two aspects. 

{\textit{Aspect~1.} When we injected non-generalizable representations of increasing magnitudes to the DNN, the first aspect was to test whether the disentangled distributions could objectively reflect both the increasing significance of non-generalizable interactions and the stability of generalizable interactions.}

Specifically, we followed the same experimental settings in Section~\ref{sec:generalization_power}, and injected non-generalizable representations of different magnitudes to the DNN with the increasing values of $\sigma^2$. Figure~\ref{fig:verify_disentangle} shows the disentangled spindle-shaped distribution of non-generalizable interactions increased along with continuously injecting increasing non-generalizable representations. In comparison, the disentangled decay-shaped distribution of generalizable interactions was not affected a lot. This experiment verified the faithfulness of our theory.

\textit{Aspect 2.} The second aspect was the matching error between the theoretical distribution and the true distribution measured in real DNNs. Specifically, Let $\textbf{A}^{(m), +}$ and $\textbf{A}^{(m), -}$ denote the real distribution of interactions extracted from a DNN.  $\textbf{A}_{\text{theory}}^{(m), +}$ and $\textbf{A}_\text{theory}^{(m), -}$ denote the theoretically estimated distribution in Equation~(\ref{eq:disentangle}). In this way, we computed the absolute matching error {\small$\textbf{A}^{(m),+}_{\text{residual}} = \vert \textbf{A}^{(m),+} - \textbf{A}_{\text{theory}}^{(m), +} \vert$} and {\small$\textbf{A}^{(m),-}_{\text{residual}} = \vert \textbf{A}^{(m),-} - \textbf{A}_{\text{theory}}^{(m), -} \vert$} {to evaluate the matching quality.}

Figure~\ref{fig:verify_disentangle} shows that strength of the matching error was small, which indicated that the disentangling algorithm effectively extracted the distribution of generalizable interactions and the distribution of non-generalizable interactions.

\begin{figure*}[t]
    \centering
    \includegraphics[width=\linewidth]
    {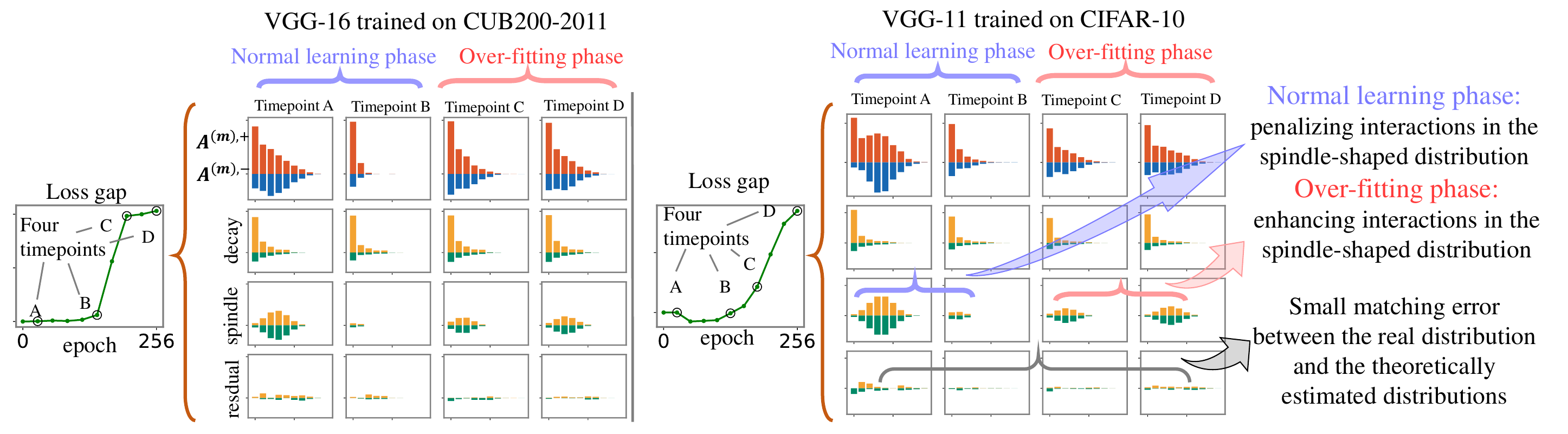}
    \caption{The distributions of generalizable interactions ($\textbf{A}^{(m),+}_{\text{decay}}, \textbf{A}^{(m),-}_{\text{decay}}$) and the distribution of non-generalizable interactions $\{\textbf{A}^{(m),+}_{\text{spindle}}, \textbf{A}^{(m),-}_{\text{spindle}}\}$ extracted from real DNNs at different timepoints of the training process. In the normal learning phase, the DNN mainly penalized interactions in a spindle-shaped distribution and learned interactions in a decay-shaped distribution. In the overfitting phase, the DNN further learned interactions in a spindle-shaped distribution. Please see Appendix~\ref{sec:more-results-fig5} for results of other two DNNs.}
    \label{fig:two_stage_disentang}
\end{figure*}

\textbf{Using our theory to disentangle the two distributions in real-world applications.} We used our theory to disentangle the distribution of the generalizable interactions and that of the non-generalizable interactions from various DNNs trained on various datasets. We used the same networks and datasets introduced in Section~\ref{sec:two_stage}.

Figure~\ref{fig:two_stage_disentang} shows the two distributions disentangled from different DNNs. Specifically, we visualized the changes of the distributions of the generalizable interactions and non-generalizable interactions in the training process.

We found that the learning phase (when the loss gap was approximately zero) mainly removed interactions following a spindle-shaped distribution, {and such interactions} were widely considered purely noises caused by the random parameter initialization. Meanwhile, interactions following a decay-shaped distribution were gradually learned by the end of the learning phase. Then, in the overfitting phase (when the loss gap increased significantly), the interactions following a spindle-shaped distribution emerged again. This indicated that the DNN began to encode non-generalizable high-order mutually offsetting interactions, which was a typical sign of overfitting.

\section{Conclusion and discussions}
{
In this paper, we have analyzed the generalization power of a DNN in terms of the generalization power of interactions encoded by the DNN. We have proposed and further experimentally verified that generalizable interactions in a DNN usually follow a decay-shaped distribution, while non-generalizable interactions usually follow a spindle-shaped distribution over different orders.  Given a DNN, we further propose a method to disentangle the two distributions of interactions encoded by the DNN. The experiments have verified that the theoretically disentangled distributions well match the true distribution of interactions in a real DNN. Moreover, various experiments have shown that the spindle-shaped distribution and the decay-shaped distribution successfully reflect the generalization power of interactions.}

{
However, our work is still far from a ultimate theory towards the explanation of the generalization power of interactions. Our findings only describe a general trend of how generalizable interactions and non-generalizable interactions distribute over different orders. Whereas, given a specific sample, our theory cannot explain a microscopic perspective. Specifically, We cannot ensure that every detail interaction in a decay-shaped distribution is generalizable, or ensure every interaction in a spindle-shaped distribution is non-generalizable.}

\section*{Impact statement}
This paper presents work whose goal is to advance the field of Machine Learning. There are many potential societal consequences of our work, none which we feel must be specifically highlighted here.


\bibliographystyle{unsrtnat}
\bibliography{references}

\newpage
\appendix
\onecolumn
\section{Related work}

Analyzing and quantifying the generalization power of deep neural networks (DNNs) is a crucial issue in deep learning. Existing explanations often either analyze the loss gap~\cite{generalization_bounds, bousquet2020sharper, deng2021toward, haghifam2020sharpened, haghifam2021towards} or focus on the smoothness of the loss landscape~\cite{flat_minima, loss_lanscape, foret2021sharpness, kwon2021asam}. Additionally, there are some researches focus on analyzing the DNN's generalization power in high-dimensional feature spaces~\citep{petrini2022learning, boopathy2023model, dyballa2024separability, nikolikj2024generalization}.

However, recent advances in interaction-based theory provide a more direct perspective to analyze the generalization power of DNNs. Specifically, the interaction-based methods define and quality the interaction effect encoded by DNNs. Since the output of a DNN can be faithfully explained as the sum of all AND-OR interactions, the generalization power of the entire DNNs can be also roughly regard as the integration of the generalization power of interactions encoded by the DNN.

\textbf{Literature in guaranteeing the faithfulness of defining and disentangling a DNN's inference patterns.} \citet{ren2023AOG} have discovered and \citet{ren2024where} have proved that, given a DNN, there always exists a logical model consisting of AND-OR inference patterns, which can using a small set of AND-OR interactions to accurately predict the DNN's outputs on all $2^n$ masked states of the input sample. Furthermore, \citet{li2023does} have shown that the salient interactions extracted from a DNN can be shared across different samples. Besides, they have also discovered that salient interactions exhibited remarkable discrimination power in classification tasks. \citet{chen2024defining} have proposed a method to extract interactions that are generalizable across different models. These findings suggest that interactions act as primitive inference patterns encoded by DNNs, forming the theoretical foundation of interaction-based theoretical frameworks.

\textbf{Literature in explaining the generalization power of DNNs from the perspective of interactions.} Recent achievements have shown that interactions play a crucial role in explaining the hidden factors affecting a DNN's adversarial robustness~\cite{ren2021towards}, adversarial transferability~\cite{wang2021unified}, and generalization power~\cite{zhou2024generalization} of a DNN. \citet{deng2022discovering} have discovered and theoretically proved the existence of a representation bottleneck in DNNs, which limits their ability to encode interactions of intermediate complexity. \citet{ren2023bayesian} have found that Bayesian neural networks (BNNs) tend to avoid encoding complex interactions, which helps explain the good adversarial robustness of BNNs. \citet{liu2023towards} have discovered that DNNs learn simple interactions more easily than complex interactions. \citet{zhang2024two} and \citet{ren2024towards} have discovered and analyzed the two-phase dynamics in DNNs' learning process.

\section{Three common conditions for sparse interactions}
\label{sec:apdx-condition-for-sparsity}

\citet{ren2024where} have formulated three mathematical conditions for the sparsity of AND interactions, as follows.

\textbf{Condition 1.} \textit{The DNN does not encode interactions higher than the $M$-th order: {\small$\forall \ S\in \{S\subseteq N \mid \vert S\vert \ge M+1\}, \ I_S^\text{AND}$}.}

Condition 1 implies that the DNN does not encode extremely high-order interactions. This is because extremely high-order interactions usually represent very complex and over-fitted patterns, which are unnecessary and unlikely to be learned by the DNN in real applications.

\textbf{Condition 2.} \textit{Let us consider the average network output {\small$\bar{u}^{(k)}\overset{\text{\rm def}}{=}\mathbb{E}_{|S|=k}[v(\boldsymbol{x}_S)-v(\boldsymbol{x}_\emptyset)]$} over all masked samples $\boldsymbol{x}_S$ with $k$ unmasked input variables. This average network output monotonically increases when $k$ increases: $\forall \ k' \le k$, we have {\small$\bar{u}^{(k')} \le \bar{u}^{(k)}$}.}

Condition 2 implies that a well-trained DNN is likely to have higher classification confidence for input samples that are less masked.

\textbf{Condition 3.} \textit{Given the average network output $\bar{u}^{(k)}$ of samples with $k$ unmasked input variables, there is a polynomial lower bound for the average network output of samples with $k' (k'\le k)$ unmasked input variables: {\small $\forall \ k' \le k, \ \bar{u}^{(k')} \ge (\frac{k'}{k})^p \ \bar{u}^{(k)}$}, where $p>0$ is a positive constant.}

Condition 3 implies that the classification confidence of the DNN does not significantly degrade
on masked input samples. The classification/detection of masked/occluded samples is common in real scenarios. In this way, a well-trained DNN usually learns to classify a masked input sample based on local information (which can be extracted from unmasked parts of the input) and thus should not yield a significantly low confidence score on masked samples.

\section{Proof of Universal Matching property (Theorem~\ref{them:matching})}
\label{proof:universal-matching}

\begin{proof} \textbf{(1) Universal matching property of AND interactions.}

We will prove that output component $v_{\text{\rm and}}(\boldsymbol{x}_S)$ on all $2^n$ masked samples $\{\boldsymbol{x}_S:S\subseteq N\}$ could be universally explained by the all interactions in $S\subseteq N$, \emph{i.e.}, $\forall \emptyset \neq S\subseteq N, v_{\text{\rm and}}(\boldsymbol{x}_S)=\sum_{\emptyset \neq T\subseteq S}I^\text{AND}_T + v(\boldsymbol{x}_\emptyset)$. In particular, we define $v_{\text{\rm and}}(\boldsymbol{x}_\emptyset)=v(\boldsymbol{x}_\emptyset)$ (\textit{i.e.}, we attribute output on an empty sample to AND interactions).

Specifically, the AND interaction is defined as $I^\text{AND}_T =  \sum\nolimits_{L \subseteq T}(-1)^{|T|-|L|}u^{\text{AND}}_L$ in Theorem~\ref{them:matching} 
To compute the sum of AND interactions $\sum_{\emptyset \neq T\subseteq S}I^\text{AND}_T =  \sum\nolimits_{\emptyset \neq T \subseteq S} \sum\nolimits_{L \subseteq T} (-1)^{\vert T \vert - \vert L \vert} u^{\text{AND}}_L$, we first exchange the order of summation of the set $L\subseteq T\subseteq S$ and the set $T \supseteq L$. 
That is, we compute all linear combinations of all sets $T$ containing $L$ with respect to the model outputs $u^{\text{AND}}_L$ given a set of input variables $L$, \textit{i.e.}, $\sum\nolimits_{T: L \subseteq T \subseteq S} (-1)^{|T|-|L|}u^{\text{AND}}_L$. Then, we compute all summations over the set $L\subseteq S$.

In this way, we can compute them separately for different cases of $L\subseteq T\subseteq S$. In the following, we consider the cases (1) $L = S = T$, and (2) $L\subseteq T\subseteq S, L\ne S$, respectively.

(1) When $L=S=T$, the linear combination of all subsets $T$ containing $L$ with respect to the model output $u^{\text{AND}}_L$ is $(-1)^{|S|-|S|} u^{\text{AND}}_L = u^{\text{AND}}_L$.

(2) When $L\subseteq T\subseteq S, L\ne S$, the linear combination of all subsets $T$ containing $L$ with respect to the model output $u^{\text{AND}}_L$ is $\sum\nolimits_{T: L \subseteq T \subseteq S} (-1)^{|T|-|L|}u^{\text{AND}}_L$. For all sets $T: S\supseteq T\supseteq L$, let us consider the linear combinations of all sets $T$ with number $|T|$ for the model output $u^{\text{AND}}_L$, respectively. Let $m := |T| - |L|$, ($0\le m\le |S|-|L|$),  then there are a total of $C_{|S|-|L|}^{m}$ combinations of all sets $T$ of order $|T|$. Thus, given $L$, accumulating the model outputs $u^{\text{AND}}_L$ corresponding to all $T\supseteq L$, then $\sum\nolimits_{T: L \subseteq T \subseteq S} (-1)^{|T|-|L|}u^{\text{AND}}_L = u^{\text{AND}}_L \cdot \underbrace{\sum\nolimits_{m=0}^{\vert S \vert - \vert L \vert} C_{|S|-|L|}^m(-1)^m}_{=0} = 0$. Please see the complete derivation of the following formula.

\begin{equation}\begin{aligned}
    &\sum\nolimits_{\emptyset \neq T \subseteq S} I_{\text{and}}(T\vert \boldsymbol{x}) \\
    = &  \sum\nolimits_{\emptyset \neq T \subseteq S} \sum\nolimits_{L \subseteq T} (-1)^{\vert T \vert - \vert L \vert} u^{\text{AND}}_L \\
    = & \sum\nolimits_{L \subseteq S} \sum\nolimits_{T: L \subseteq T \subseteq S} (-1)^{\vert T \vert - \vert L \vert} u^{\text{AND}}_L  - u^{\text{AND}}_\emptyset \\
    = & \underbrace{u^\text{AND}_S}_{L = S} + \sum\nolimits_{L \subseteq S, L \neq S} u^{\text{AND}}_L \cdot \underbrace{\sum\nolimits_{m=0}^{\vert S \vert - \vert L \vert} C_{|S|-|L|}^m(-1)^m}_{=0}   - u^{\text{AND}}_\emptyset \\
     = & u^\text{AND}_S  - u^{\text{AND}}_\emptyset \\
     = & u^\text{AND}_S  - u_\emptyset
\end{aligned}\end{equation}

Thus, we have $\forall \emptyset \neq S\subseteq N, v_{\text{\rm and}}(\boldsymbol{x}_S)=\sum_{\emptyset \neq T\subseteq S}I^\text{AND}_T + v(\boldsymbol{x}_\emptyset)$.

\textbf{(2) Universal matching property of OR interactions.}

{According to the definition of OR interactions, we will derive that $\forall S\subseteq N, u^\text{OR}_S=\sum_{T:T\cap S\neq \emptyset}I^\text{OR}_S$, 
where we define $u^\text{OR}_\emptyset=0$ (recall that in Step (1), we attribute the output on empty input to AND interactions).}

{Specifically, the OR interaction is defined as $I^\text{OR}_S =  -\sum\nolimits_{L \subseteq T}(-1)^{|T|-|L|}u^\text{OR}_{N \setminus L}$ in Theorem~\ref{them:matching}
Similar to the above derivation of the Universal matching property of AND interactions, to compute the sum of OR interactions $\sum\nolimits_{T:T \cap S \neq \emptyset} I^\text{OR}_T = \sum\nolimits_{T:T \cap S \neq \emptyset} \left[- \sum\nolimits_{L \subseteq T} (-1)^{\vert T \vert - \vert L \vert} u^\text{OR}_{N \setminus L} \right]$, we first exchange the order of summation of the set $L\subseteq T \subseteq N$ and the set $T:T \cap S \neq \emptyset$. That is, we compute all linear combinations of all sets $T$ containing $L$ with respect to the model outputs $u^\text{OR}_{N \setminus L}$ given a set of input variables $L$, \textit{i.e.}, $\sum\nolimits_{T: T \cap S \neq \emptyset, T \supseteq L} (-1)^{\vert T \vert - \vert L \vert} u^\text{OR}_{N \setminus L}$. Then, we compute all summations over the set $L\subseteq N$.}

{In this way, we can compute them separately for different cases of $L\subseteq T\subseteq N, T \cap S \neq \emptyset$. In the following, we consider the cases (1) $L = N \setminus S$, (2) $L=N$, (3) $L \cap S \neq \emptyset, L \neq N$, and (4) $L \cap S=\emptyset, L \neq N \setminus S$, respectively.}

{(1) When $L = N \setminus S$, the linear combination of all subsets $T$ containing $L$ with respect to the model output $u^\text{OR}_{N \setminus L}$ is $\sum\nolimits_{T: T \cap S \neq \emptyset, T \supseteq L} (-1)^{\vert T \vert - \vert L \vert} u^\text{OR}_{N \setminus L}= \sum\nolimits_{T: T \cap S \neq \emptyset, T \supseteq L} (-1)^{\vert T \vert - \vert L \vert} u^\text{OR}_S$. For all sets $T: T\supseteq L, T \cap S \neq \emptyset$ (then $T \neq N \setminus S, T \neq L$), let us consider the linear combinations of all sets $T$ with number $|T|$ for the model output $u^\text{OR}_S$, respectively. Let $|T'| := |T| - |L|$, ($1\le |T'|\le |S|$),  then there are a total of $C_{|S|}^{|T'|}$ combinations of all sets $T'$ of order $|T'|$. 
Thus, given $L$, accumulating the model outputs $u^\text{OR}_S$ corresponding to all $T\supseteq L$, then $\sum\nolimits_{T: T \cap S \neq \emptyset, T \supseteq L} (-1)^{\vert T \vert - \vert L \vert} u^\text{OR}_{N \setminus L} = u^\text{OR}_S \cdot \underbrace{\sum\nolimits_{|T'|=1}^{\vert S \vert } C_{|S|}^{|T'|}(-1)^{|T'|}}_{=-1} = -u^\text{OR}_S$.}

{(2) When $L=N$ (then $T=N$), the linear combination of all subsets $T$ containing $L$ with respect to the model output $u^\text{OR}_{N \setminus L}$ is $\sum\nolimits_{T: T \cap S \neq \emptyset, T \supseteq L} (-1)^{\vert T \vert - \vert L \vert} u^\text{OR}_{N \setminus L}= (-1)^{\vert N \vert - \vert N \vert} u^\text{OR}_\emptyset = u^\text{OR}_\emptyset$.}

{(3) When $L \cap S \neq \emptyset, L \neq N$, the linear combination of all subsets $T$ containing $L$ with respect to the model output $u^\text{OR}_{N \setminus L}$ is $\sum\nolimits_{T: T \cap S \neq \emptyset, T \supseteq L} (-1)^{\vert T \vert - \vert L \vert} u^\text{OR}_{N \setminus L}$. For all sets $T: T\supseteq L, T \cap S \neq \emptyset$, let us consider the linear combinations of all sets $T$ with number $|T|$ for the model output $u^\text{OR}_S$, respectively. Let us split $|T| - |L|$ into $|T'|$ and $|T''|$, \textit{i.e.},$|T| - |L| = |T'| + |T''|$, where $T'=\{i|i\in T, i\notin L, i\in N\setminus S\}$, $T''=\{i|i\in T, i\notin L, i\in S\}$ (then $0\le|T''|\le|S|-|S\cap L|$) and $|T'| + |T''| + |L| = |T|$. In this way, there are a total of $C_{|S|-|S\cap L|}^{|T''|}$ combinations of all sets $T''$ of order $|T''|$. Thus, given $L$, accumulating the model outputs $v_{\text{or}}(\boldsymbol{x}_{N\setminus L})$ corresponding to all $T\supseteq L$, then $\sum\nolimits_{T: T \cap S \neq \emptyset, T \supseteq L} (-1)^{\vert T \vert - \vert L \vert} u^\text{OR}_{N \setminus L} = u^\text{OR}_{N \setminus L} \cdot \sum_{T' \subseteq N\setminus S \setminus L}\underbrace{\sum\nolimits_{\vert T'' \vert = 0}^{\vert S \vert-\vert S \cap L \vert} C_{\vert S \vert - \vert S \cap L \vert}^{\vert T''\vert } (-1)^{\vert T' \vert + \vert T'' \vert}  }_{=0} = 0$.}

{(4) When $L \cap S=\emptyset, L \neq N \setminus S$, the linear combination of all subsets $T$ containing $L$ with respect to the model output $u^\text{OR}_{N \setminus L}$ is $\sum\nolimits_{T: T \cap S \neq \emptyset, T \supseteq L} (-1)^{\vert T \vert - \vert L \vert} u^\text{OR}_{N \setminus L}$. Similarly, let us split $|T| - |L|$ into $|T'|$ and $|T''|$, \textit{i.e.},$|T| - |L| = |T'| + |T''|$, where $T'=\{i|i\in T, i\notin L, i\in N\setminus S\}$, $T''=\{i|i\in T, i\in S\}$ (then $0\le|T''|\le|S|$) and $|T'| + |T''| + |L| = |T|$. In this way, there are a total of $C_{|S|}^{|T''|}$ combinations of all sets $T''$ of order $|T''|$. Thus, given $L$, accumulating the model outputs $v_{\text{or}}(\boldsymbol{x}_{N\setminus L})$ corresponding to all $T\supseteq L$, then $\sum\nolimits_{T: T \cap S \neq \emptyset, T \supseteq L} (-1)^{\vert T \vert - \vert L \vert} u^\text{OR}_{N \setminus L} = u^\text{OR}_{N \setminus L} \cdot \sum_{T' \subseteq N\setminus S \setminus L}\underbrace{\sum\nolimits_{\vert T'' \vert = 0}^{\vert S \vert} C_{\vert S \vert }^{\vert T''\vert } (-1)^{\vert T' \vert + \vert T'' \vert}  }_{=0} = 0$.}

{Please see the complete derivation of the following formula.}
\begin{equation}
\begin{aligned}
\sum\nolimits_{T:T \cap S \neq \emptyset} I^\text{OR}_T
        &= \sum\nolimits_{T:T \cap S \neq \emptyset} \left[- \sum\nolimits_{L \subseteq T} (-1)^{\vert T \vert - \vert L \vert} u^\text{OR}_{N \setminus L} \right]\\
        &= - \sum\nolimits_{L \subseteq N} \sum\nolimits_{T: T \cap S \neq \emptyset, T \supseteq L} (-1)^{\vert T \vert - \vert L \vert} u^\text{OR}_{N \setminus L} \\
        &=  - \left[\sum_{\vert T' \vert = 1}^{\vert S \vert} C_{\vert S \vert}^{\vert T' \vert} (-1)^{\vert T' \vert} \right] \cdot \underbrace{u^\text{OR}_S}_{L=N\setminus S} - \underbrace{u^\text{OR}_\emptyset}_{L=N} \\
        &\quad- \sum_{L \cap S \neq \emptyset, L \neq N} \left[\sum_{T' \subseteq N\setminus S \setminus L} \left( \sum_{\vert T'' \vert = 0}^{\vert S \vert-\vert S \cap L \vert} C_{\vert S \vert - \vert S \cap L \vert}^{\vert T''\vert } (-1)^{\vert T' \vert + \vert T'' \vert} \right) \right]\cdot u^\text{OR}_{N \setminus L}  \\
        &\quad- \sum_{L \cap S=\emptyset, L \neq N \setminus S} \left[ \sum_{T' \subseteq N\setminus S \setminus L} \left( \sum_{\vert T'' \vert=0}^{\vert S \vert} C_{\vert S \vert}^{\vert T'' \vert} (-1)^{\vert T' \vert + \vert T'' \vert}\right) \right] \cdot u^\text{OR}_{N \setminus L}  \\
        &=  - (-1) \cdot u^\text{OR}_S - u^\text{OR}_\emptyset - \sum_{L \cap S \neq \emptyset, L \neq N} \left[\sum_{T' \subseteq N\setminus S \setminus L} 0 \right]\cdot u^\text{OR}_{N \setminus L}  \\
        &\quad- \sum_{L \cap S=\emptyset, L \neq N \setminus S}\left[\sum_{T' \subseteq N\setminus S \setminus L} 0 \right] \cdot u^\text{OR}_{N \setminus L}  \\
        &= u^\text{OR}_S - u^\text{OR}_\emptyset\\
        &= u^\text{OR}_S
\end{aligned}
\end{equation}

\textbf{(3) Universal matching property of AND-OR interactions.}

With the Universal matching property of AND interactions and the Universal matching property of OR interactions, we can easily get {$v(\boldsymbol{x}_S) =  u^\text{AND}_S  + u^\text{OR}_S 
=v(\boldsymbol{x}_\emptyset) + \sum_{\emptyset \neq T\subseteq S}I^\text{AND}_T +\!\! \sum_{T: T\cap S \neq \emptyset}I^\text{OR}_T$}, thus, we obtain the Universal matching property of AND-OR interactions.

\end{proof}

\section{Details of optimizing $\{\gamma_T\}$ to extract the sparsest AND-OR interactions}
\label{sec:apdx-optimize-pq}
A method is proposed~\cite{li2023defining, chen2024defining} to simultaneously extract AND interactions $I_S^\text{AND}$
and OR interactions $I_S^\text{OR}$ from the network output. Given a  masked sample $\boldsymbol{x}_T$, \cite{li2023defining} proposed to learn a decomposition $v(\boldsymbol{x}_T)=u_T^\text{AND} + u_T^\text{OR}$ towards the sparsest interactions. 
The component {$u_T^\text{AND}$} was explained by AND interactions, and the component {$u_T^\text{OR}$} was explained by OR interactions.
Specifically, they decomposed $v(\boldsymbol{x}_T)$ into $u_T^\text{AND}= 0.5  \ v(\boldsymbol{x}_T)+\gamma_T$ and $u_T^\text{AND}= 0.5 \cdot v(\boldsymbol{x}_T) -\gamma_T$, where $\{\gamma_T:T\subseteq N\}$ is a set of learnable variables that determine the decomposition. In this way, the AND interactions and OR interactions can be computed according to Theorem.~\ref{them:matching}, \textit{i.e.}, $I_{\text{and}}(S|\boldsymbol{x})=\sum\nolimits_{T \subseteq S}(-1)^{|S|-|T|} v^\text{AND}_T)$, and  $I_{\text{or}}(S | \boldsymbol{x})=-\sum\nolimits_{T \subseteq S}(-1)^{|S|-|T|} u^\text{OR}_{N \setminus T})$.

The parameters $\{\gamma_T\}$ were learned by minimizing the following LASSO-like loss to obtain sparse interactions:
\begin{equation}
\label{eq:loss-pq}
    \min_{\{\gamma_T\}} \sum_{S\subseteq N} \vert I^\text{AND}_S \vert + \vert I^\text{OR}_S \vert
\end{equation}

\textbf{Removing small noises.} A small noise $\delta_S$ in the network output may significantly affect the extracted interactions, especially for high-order interactions. Thus, ~\cite{li2023defining} proposed to learn to remove a small noise term $\delta_S$ from the computation of AND-OR interactions.
Specifically, the decomposition was rewritten as {$u_T^\text{AND}=0.5 (v(\boldsymbol{x}_T) -\delta_T) +\gamma_T$} and {$u_T^\text{OR}=0.5 (v(\boldsymbol{x}_T)-\delta_T) +\gamma_T$}.
Thus, the parameters {$\{\delta_T\}$,} and {$\{\gamma_T\}$} are simultaneously learned by minimizing the loss function in Eq.~(\ref{eq:loss-pq}).
The values of {$\{\delta_T\}$} were constrained in $[-\zeta, \zeta]$ where $\zeta=0.02\cdot \mathbb{E}_{\boldsymbol{x}} \vert v(\boldsymbol{x})-v(\boldsymbol{x}_\emptyset) \vert$.


\section{Discussion: overfitting towards very few samples may cause noises.}
\label{sec:apdx-discuss-adding-noise}
The overfitted feature representation or an overfitted interaction exhibit clear utility on very few samples (or even a single), rather than on a large, broad sample. However, as the cost of learning such interactions, such the additionally learned features usually cause the DNN generate a lot of non-discriminative noise activations, which is modeled as noise on the input or neural network parameters.

\section{Towards the Two-Stage Problem of Bert-Medium in Figure~\ref{fig:stage}}
\label{sec:apdx-bert-medium}
\citet{zhang2024two} proposed that DNNs with entirely random weights mainly encode medium-order and high-order interactions under certain assumptions. Their experiments further confirmed that most randomly initialized DNNs align well with this proposition. However, this may not fully apply to Bert-Medium.

In the first phase, Bert-Medium primarily encodes extremely low-order interactions, which is not consistent with the proposition. In the second phase, it generally learns high-order interactions, which is consistent with the proposition. This may come from the fact that Bert-Medium is pre-trained on a large corpus, which may have already encoded low-order interactions. However, this inconsistency does not significantly affect our theory. Our method still extracted the two distributions of interactions in the overfitting phase of Bert-Medium.
\begin{figure*}[ht]
    \centering
    \includegraphics[width=\linewidth]{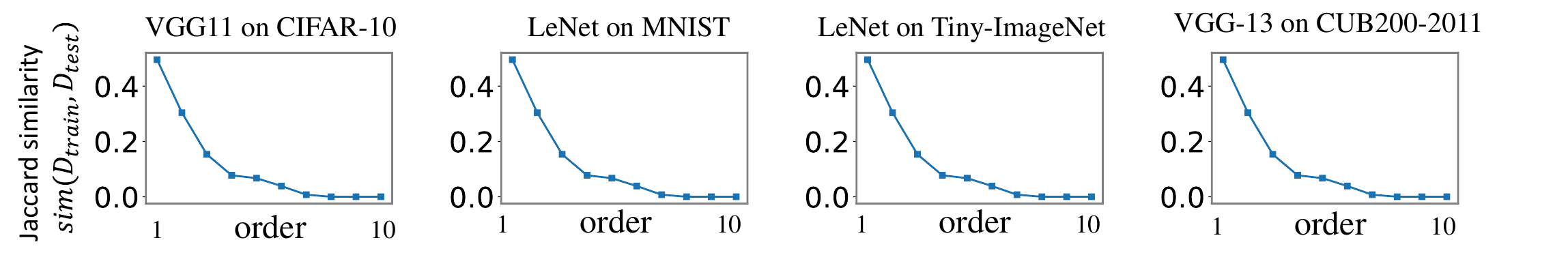}
    \caption{{The averaged Jaccard similarity between interactions extracted from training samples and those extracted from testing samples across the first 10 categories in the datasets. low-order interactions have a higher value of $E_c Sim(D_{\text{train}}, D_{\text{test}})$ than high-order interactions.}}
    \label{fig:generalization_order}
\end{figure*}
\section{Experiments verification of Claim~\ref{claim:generalization}}
\label{sec:apdx-verify-generalization}
Define\ref{def:generalization} shows a typical definition of the generalization power of interactions proposed by \citet{zhou2024generalization}. If an interaction, which frequently appears on the training samples, also frequently appears on the testing samples, then this interaction is considered to be generalized to testing samples; otherwise, not. In this way, they have found the following empirical property of interactions of each order. The low-order interactions have a stronger generalization power than high-order interactions. We conducted the following experiments to verify Claim~\ref{claim:generalization}.

We computed the average Jaccard similarity $E_c Sim(D_{\text{train}}, D_{\text{test}})$(introduced in Define\ref{def:generalization}) between interactions extracted from training samples and those extracted from testing samples across 10 categories in the datasets. We trained different DNNs, including LeNet\citep{lecun1998gradient}, VGG-11, VGG-13, and AlexNet, on different datasets, including MNIST, CIFAR-10, CUB200-2011, and Tiny-ImageNet. 

Figure~\ref{fig:generalization_order} shows that the Jaccard similarity of the interactions kept decreasing as the order of the interactions increased. Thus, it verified that The generalization power of high-order(complex) interactions is lower than that of low-order (simple) interactions,

\section{Experiments verification of Theorem~\ref{theorem:overfitting} }
\label{sec:apdx-exp-overfitting}
\textbf{The formulation of the matrix $M(\delta)$ in Theorem~\ref{theorem:overfitting}.} According to \citep{ren2024towards}, the matrix $M(\delta)$ is calculated as $M(\delta) =({\boldsymbol{J}}^\top \boldsymbol{J} + 2^n{\rm diag}(\boldsymbol{c}))^{-1}{\boldsymbol{J}}^\top \boldsymbol{J}$, where {\small$\boldsymbol{J} \overset{\text{\rm def}}{=} [\boldsymbol{J}(\boldsymbol{x}_{S_1}),\boldsymbol{J}(\boldsymbol{x}_{S_2}), \cdots, \boldsymbol{J}(\boldsymbol{x}_{S_{2^n}})]^\top \in \mathbb{R}^{2^n\times 2^n}$} is a matrix to represent the triggering values of $2^n$ interactions (\textit{w.r.t.} $2^n$ columns) on $2^n$ masked samples (\textit{w.r.t.} $2^n$ rows). {\small$\boldsymbol{x}_{S_1}, \boldsymbol{x}_{S_2},\cdots,\boldsymbol{x}_{S_{2^n}}$} enumerate all masked samples. {\small$\boldsymbol{c} \overset{\text{\rm def}}{=} {\rm vec}(\{{\rm Var}[\epsilon_T]:{T\subseteq N}\})= {\rm vec}(\{2^{|T|}\sigma^2:{T\subseteq N}\})\in\mathbb{R}^{2^n}$} denotes the vector of variances of the triggering values of $2^n$ interactions. 

\begin{figure*}[ht]
    \centering
    \includegraphics[width=0.9\linewidth]{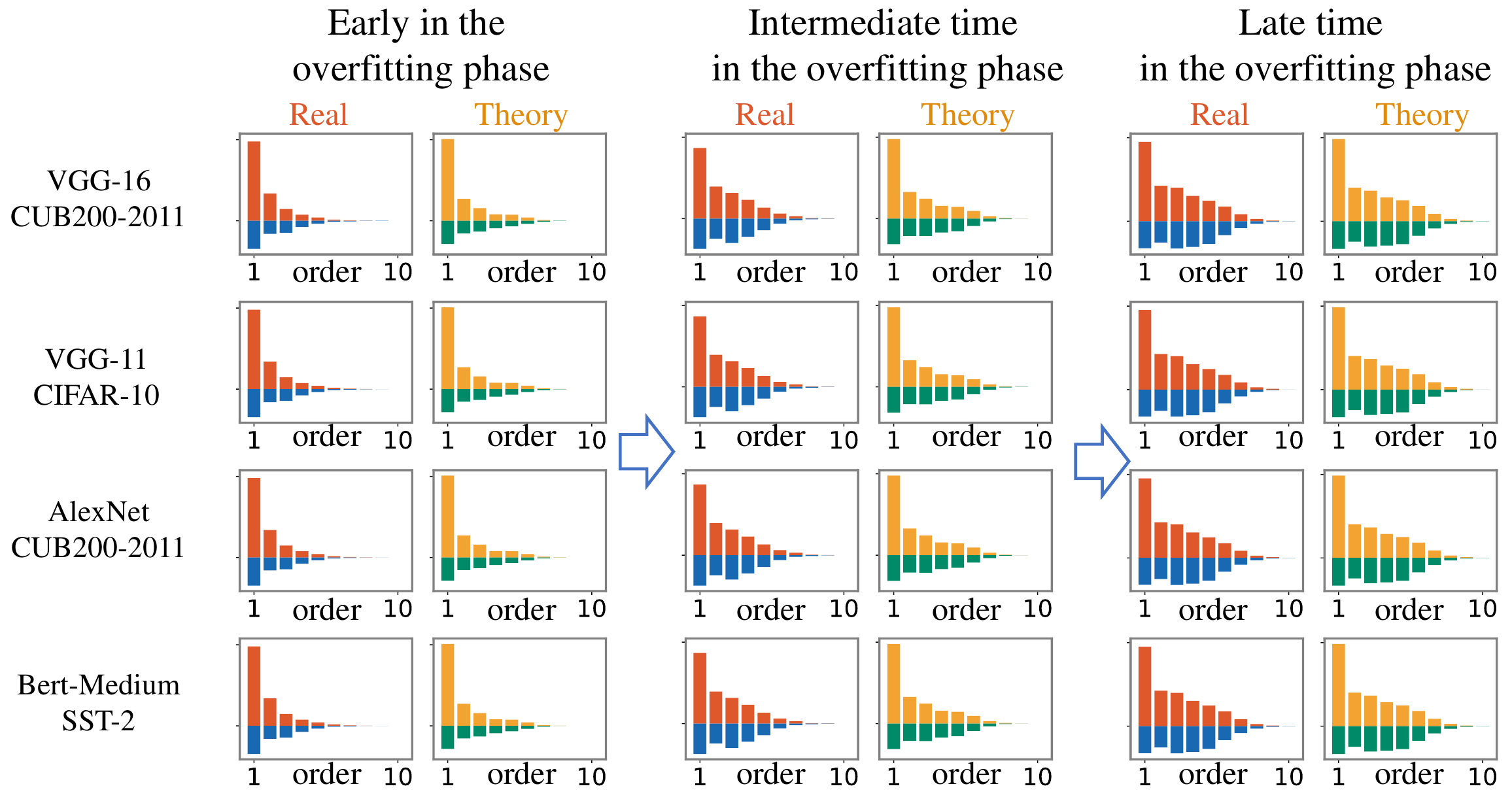}
    \caption{{Comparison between the theoretical distribution of interactions $\textbf{A}^{(m), +}_\text{decay}, \textbf{A}^{(m), -}_\text{decay}$ and the real distribution of interaction strength $\textbf{A}^{(m), +}, \textbf{A}^{(m), -}$ in the overfitting phase.}}
    \label{fig:verify_decay}
\end{figure*}

Theorem~\ref{theorem:overfitting} shows that the ubiquitous uncertainty in network parameters and training data will eliminate the mutually counteracting high-order interactions, thereby avoiding. We conducted the following experiments to verify the Theorem~\ref{theorem:overfitting}.
Specifically, We trained VGG-11, VGG-16, AlexNet on the CIFAR-10 datasets, CUB200-2011 datasets and Tiny-ImageNet datasets, respectively. We also trained BERT-Medium models on the SST-2 datasets. We calculated the distributions $\text{A}^{(m), +}, \text{A}^{(m), -}$ of interactions extracted form the three timepoints in the DNN's overfitting phase. Given the extremely overfitting DNN, we used the Theorem~\ref{theorem:overfitting} to predict the dynamics os interactions of different orders before the overfitting phase. We verified whether the Theorem~\ref{theorem:overfitting} could well predict the real dynamics of interaction strength of different orders in real DNNs.

Figure \ref{fig:verify_decay} shows that the theoretical distribution $\textbf{A}^{(m), +}_\text{decay}, \textbf{A}^{(m), -}_\text{decay}$ of interactions well matched the real distribution $\textbf{A}^{(m), +}, \textbf{A}^{(m), -}$ of interactions in the overfitting phase. This verifies that the Theorem~\ref{theorem:overfitting} could well predict the dynamics of interactions of different orders in real DNNs. In this way, we could remove the eliminate the mutually counteracting high-order interactions by setting a enough large uncertainty level $\delta$ (In our experiments, we set $\delta\leq 1e-3$).

\section{Experimental details}
\subsection{Models and datasets}
\label{sec:apdx-model-dataset}
We trained various DNNs on different datasets. Specifically, for image data, we trained ResNet-20, LeNet on the MNIST dataset (Creative Commons Attribution-Share Alike 3.0 license), VGG-11/VGG-16 on the CIFAR-10 dataset (MIT license), AlexNet/VGG-13/VGG-16 on the CUB-200-2011 dataset (license unknown). For natural language data, we trained BERT-Medium on the SST-2 dataset (license unknown). 

For the CUB-200-2011 dataset, we cropped the images to remove the background regions, using the bounding box provided by the dataset. These cropped images were resized to 224$\times$224 and fed into the DNN. For the Tiny ImageNet dataset, due to the computational cost, we selected 50 classes from the total 200 classes at equal intervals (\textit{i.e.}, the 4th, 8th,..., 196th, 200th classes). All these images were resized to 224$\times$224. For the MNIST dataset, all images were resized to 32$\times$32 for classification. 

To better demonstrate that the learning of higher-order interactions in the second phase was closely related to overfitting, we added a small ratio of label noise to the CIFAR-10 dataset, and the CUB-200-2011 dataset to boost the significance of overfitting of the DNNs. Specifically, we randomly selected 1\% training samples in the MNIST dataset and the CIFAR-10 dataset, and randomly reset their labels. We randomly selected 5\% training samples in the CUB-200-2011 dataset and randomly reset their labels.

\subsection{Details on computing interactions}
\label{sec:apdx-detail-compute-interaction}
First, we provide a summary of the mathematical settings of the hyper-parameters for interactions in Table \ref{tab:mathematical-setting}, including the scalar output function of the DNN $v(\cdot)$, the baseline value $\boldsymbol{b}$ for masking, and the threshold $\tau$. These settings are uniformly applied to all DNNs. More detailed settings for different datasets can be found below.

\textbf{Image data.} For image data, we considered image patches as input variables to the DNN. 
To generate a masked sample $\boldsymbol{x}_S$, we followed~\cite{zhang2024two} to mask the patch on the intermediate-layer feature map corresponding to each image patch in the set $N\setminus S$.
Specifically, we considered the feature map after the second ReLU layer for VGG-11/VGG-16 and the feature map after the first ReLU layer for AlexNet.  
For the VGG models and the AlexNet model, we uniformly partitioned the feature map into 8$\times$8 patches, randomly selected 10 patches from the central 6$\times$6 region (\textit{i.e.}, we did not select patches that were on the edges), and considered each of the 10 patches as an input variable in the set $N$ to calculate interactions. 
We considered each of the 10 patches as an input variable in the set $N$ to calculate interactions.
We used a zero baseline value to mask the input variables
in the set $N\setminus S$ to obtain the masked sample $\boldsymbol{x}_S$.

\textbf{Natural language data.} We considered the input tokens as input variables for each input sentence. Specifically, we randomly selected 10 words that are meaningful (\textit{i.e.}, not including stopwords, special characters, and punctuations) as input variables in the set $N$ to calculate interactions. We used the “mask” token with the token id 103 to mask the tokens in the set $N\setminus S$ to obtain the masked sample $\boldsymbol{x}_S$.

For all DNNs and datasets, we randomly selected 50 samples from the testing set to compute interactions, and averaged the interaction strength of the $k$-th order on each sample to obtain {\small$I^{(k)}_{\text{real}}$}.

\subsection{Normalization methods}
\label{sec:apdx-normalization}
In the experiment of analyzing the distribution of interactions at the different timepoints in the training process of DNNs, we need to normalize the interaction strength of the $k$-th order on each sample. Specifically, for interactions of each $k$-th order, we normalized the strength of salient interactions as $\tilde{I}^\text{AND} = I^\text{AND} / Z$ and $\tilde{I}^\text{OR} = I^\text{OR} / Z$, where $Z$ is the averaged sum of the absolute values of all salient interactions of the $1$-order across the all samples, $\textit{i.e.}, Z = E_x \sum_{S\in \Omega^\text{AND}, \vert S \vert = 1} |I^\text{AND}_S| + E_x \sum_{S\in \Omega^\text{OR}, \vert S \vert = 1} |I^\text{OR}_S|$. The normalization removes the effect of the explosion of output values during the training process and enables us to only analyze the relative distribution of interaction strength.

\subsection{Compute resources}
All DNNs can be trained within 12 hours on a single NVIDIA GeForce RTX 3090 GPU (with 24G GPU memory). Computing all interactions on a single input sample usually takes 35-40 seconds, which is acceptable in real applications.

\begin{table}[t]
    \centering
    \begin{tabular}{|c|l|}
        \hline
        Output function $v(\cdot)$ & {\small$v(\boldsymbol{x})=\log \frac{p(y^{\text{truth}}|\boldsymbol{x})}{1-p(y^{\text{truth}}|\boldsymbol{x})}$} \\
        \hline        
        Threshold $\tau$ & {\small$\tau \!=\! 0.02\ \mathbb{E}_{\boldsymbol{x}} [|v(\boldsymbol{x})-v(\boldsymbol{x}_\emptyset)|]$} \\        
        \hline
        \multirow{2}{*}{Baseline value $\boldsymbol{b}$} & Image data: using the zero baseline on the feature map after ReLU \\
        \cline{2-2}
                                    & Text data: using the [MASK] token \\
        \hline
    \end{tabular}
    \caption{Mathematical setting of hyper-parameters for interactions.}
    \label{tab:mathematical-setting}
\end{table}
\newpage
\section{More experimental results}
\subsection{More results for the noise injections experiments in Section~\ref{sec:experiments}}
\label{sec:more-results-fig4}
\begin{figure*}[ht]
    \centering
    \includegraphics[width=\linewidth]{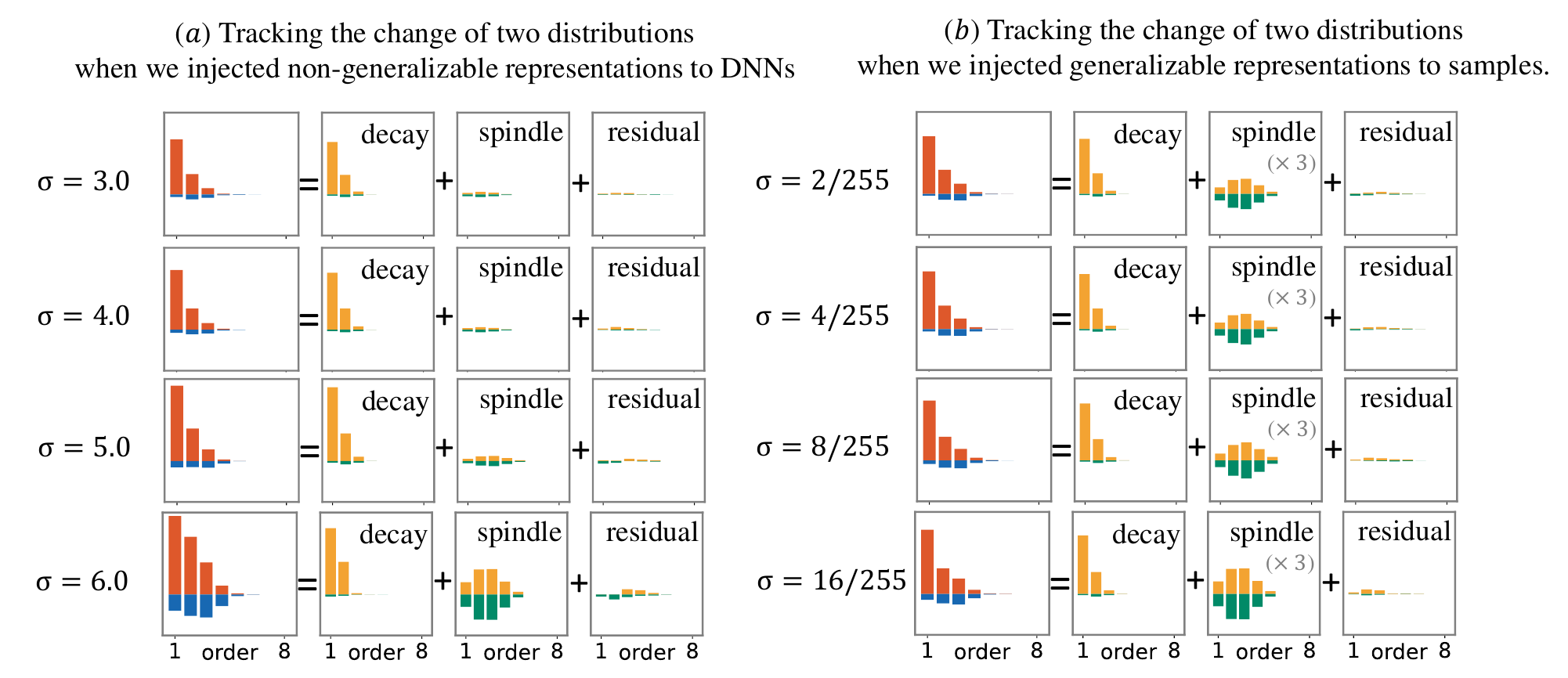}
    \caption{The decay-shaped distribution of generalizable interactions ({\small$\textbf{A}^{(m),+}_{\text{decay}},\textbf{A}^{(m),-}_{\text{decay}}$}) and the spindle-shaped distribution of non-generalizable interactions ({\small$\textbf{A}^{(m),+}_{\text{spindle}}, \textbf{A}^{(m),-}_{\text{spindle}}$}) disentangled by our method. The significance of interactions in the spindle-shaped distribution increased when we injected more non-generalizable representations to the DNN, but the significance of interactions in a decay-shaped distribution was not affected much. It verified the faithfulness of our method. }
    \label{fig:more_verify_disentangle}
\end{figure*}

\subsection{More results for the disentanglement of the real DNNs in Section~\ref{sec:experiments}}
\label{sec:more-results-fig5}

\begin{figure*}[ht]
    \centering
    \includegraphics[width=\linewidth]{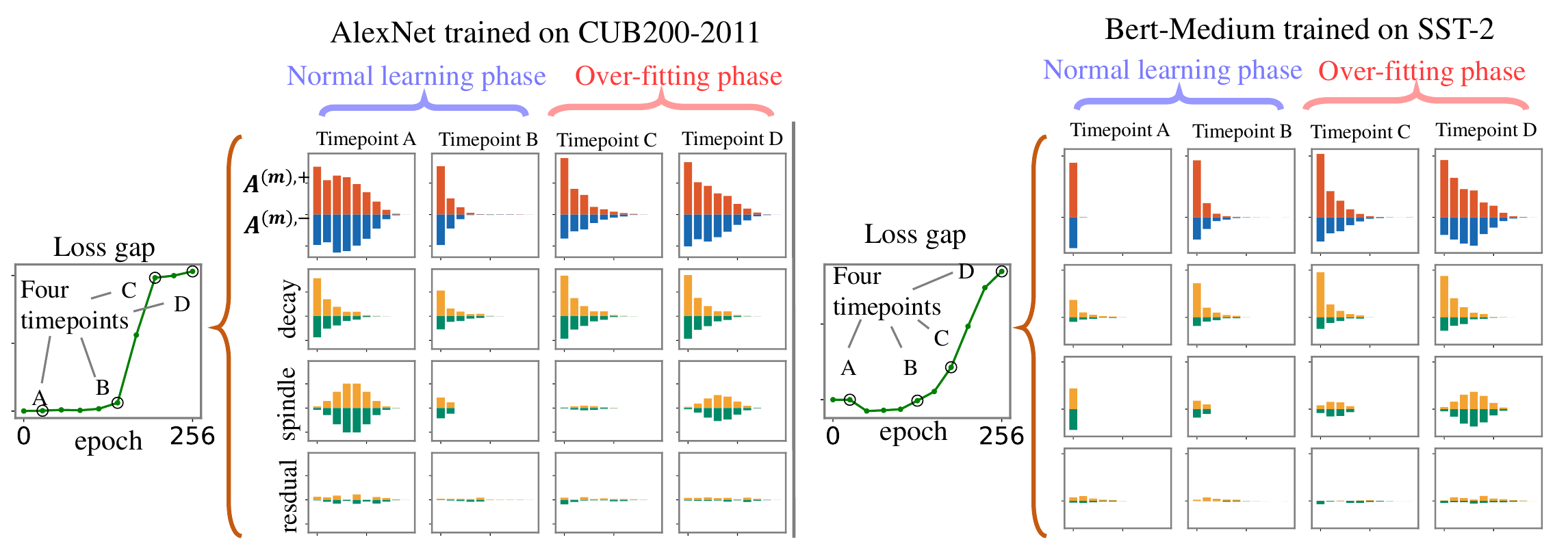}
    \caption{The distributions of generalizable interactions ($\textbf{A}^{(m),+}_{\text{decay}}, \textbf{A}^{(m),-}_{\text{decay}}$) and the distribution of non-generalizable interactions $\{\textbf{A}^{(m),+}_{\text{spindle}}, \textbf{A}^{(m),-}_{\text{spindle}}\}$ extracted from real DNNs at different timepoints of the training process. In the normal learning phase, the DNN mainly penalized interactions in a spindle-shaped distribution and learned interactions in a decay-shaped distribution. In the overfitting phase, the DNN further learned interactions in a spindle-shaped distribution.}
    \label{fig:more_two_stage_disentang}
\end{figure*}
\end{document}